\documentclass[runningheads]{llncs}
\usepackage{cite}
\usepackage{amsmath}
\usepackage{amssymb}
\usepackage{amsfonts}
\usepackage{algorithmic}
\usepackage{graphicx}
\usepackage{textcomp}
\usepackage{xcolor}
\usepackage{caption}
\usepackage{subcaption}
\usepackage{wrapfig} 
\usepackage{float}
\usepackage{url}

\begin{document}
\title{Comparing NARS and Reinforcement Learning: An Analysis of ONA and $Q$-Learning Algorithms}
\titlerunning{An Analysis of ONA and $Q$-Learning Algorithms}

%
\author{Ali Beikmohammadi
\and
Sindri Magnússon
}
%
\authorrunning{A. Beikmohammadi et al.}
%
\institute{Department of Computer and Systems Sciences, \\
  Stockholm University, SE-106 91 Stockholm, Sweden \\
\email{\{beikmohammadi, sindri.magnusson\}@dsv.su.se}
}

\maketitle              
\begin{sloppypar}
\begin{abstract}
In recent years, reinforcement learning (RL) has emerged as a popular approach for solving sequence-based tasks in machine learning. However, finding suitable alternatives to RL remains an exciting and innovative research area. One such alternative that has garnered attention is the Non-Axiomatic Reasoning System (NARS), which is a general-purpose cognitive reasoning framework.
In this paper, we delve into the potential of NARS as a substitute for RL in solving sequence-based tasks. To investigate this, we conduct a comparative analysis of the performance of ONA as an implementation of NARS and $Q$-Learning in various environments that were created using the Open AI gym. The environments have different difficulty levels, ranging from simple to complex.
Our results demonstrate that NARS is a promising alternative to RL, with competitive performance in diverse environments, particularly in non-deterministic ones. 

\end{abstract}
\section{Introduction} 

Reinforcement Learning (RL) is a type of machine learning that enables agents to make decisions in an environment to maximize their cumulative reward over time. Combining RL with high-capacity function approximations in model-free algorithms offers the potential to automate a wide range of decision-making and control tasks \cite{sutton2018}. Such algorithms have successfully tackled complex problems in various domains, such as game playing \cite{silver2017}, financial markets \cite{fischer2018}, robotic control \cite{kober2013}, and healthcare \cite{10.1145/3477600}.
However, RL faces challenges in environments where it is difficult or costly to generate large amount of data. This is due to the lack of compositional representations that would enable efficient learning \cite{hammer2021autonomy}.

Taking a broader perspective, the ultimate goal of Artificial General Intelligence (AGI) is to create intelligent systems that can adapt and learn to solve a broad range of tasks in diverse environments. RL has been a popular approach in this pursuit, but its limitations in handling environments with limited data and complex, abstract representations have hindered its progress towards AGI. To overcome these limitations, it is essential to explore alternative approaches that can facilitate data-efficient learning and deal with compositional representations effectively. 
One such approach is Non-Axiomatic Reasoning System (NARS), which is a promising approach that addresses the challenges posed by complex and uncertain environments, as it is a general-purpose reasoner that adapts under the Assumption of Insufficient Knowledge and Resources (AIKR) \cite{hammer2020reasoning, wang2009insufficient, wang2013non}.

Implementations based on  non-axiomatic logic have been developed, such as OpenNARS \cite{hammer2016opennars} and ONA (OpenNARS for Applications) \cite{hammer2020opennars}. ONA surpasses OpenNARS in terms of reasoning performance and has recently been compared 
with RL \cite{eberding2020sage, hammer2021autonomy}. 
Several challenges arise when comparing the performance of ONA and $Q$-Learning, a basic approach in RL \cite{watkins1992q}, algorithms. These challenges have been discussed in \cite{hammer2021autonomy} and include dealing with \textit{statements instead of states, unobservable information, one action in each step, multiple objectives, hierarchical abstraction, changing objectives, and goal achievement as reward}. 

In \cite{hammer2021autonomy}, three simple environments; Space invaders, Pong, and grid robot were used to compare ONA with $Q$-Learning \cite{watkins1992q}, and the results showed that ONA provided more stable outcomes while maintaining almost identical success ratio performance as $Q$-Learning. To enable a meaningful and fair comparison, an extra \textit{nothing} action added to the $Q$-Learner in each example since the competitor, ONA, does not assume that in every step, an action has to be chosen. However, this change raises concerns about preserving the problems' authenticity.

In this paper, we aim to investigate the potential of NARS as a substitute for RL algorithms and explore its capability to facilitate more efficient and effective learning in AGI systems. Specifically, we compare the performance of ONA and $Q$-Learning on several more challenging tasks compared to \cite{hammer2021autonomy}, including non-deterministic environments. Also, in contrast with \cite{hammer2021autonomy}, we propose selecting a random action when ONA does not recommend any action to be taken to keep the originality of the tasks/environments as much as possible. This approach can also benefit the agent in terms of exploring the environment.
Our findings provide insights into the potential of NARS as an alternative to RL algorithms for developing more intelligent and adaptive systems.

This paper is organized as follows: Methods are described in Section \ref{sectionM}, tasks and setups are expressed in Section \ref{section2}, experimental results and analyses are reported in Section \ref{section3}, and we conclude and discuss future work in Section \ref{sectionC}.

\section{Methods} \label{sectionM} 

\subsection{RL and Tabular $Q$-Learning}\label{sectionRL}
\textbf{RL}, in which an agent interacts with an unknown environment, typically is modeled as a Markov decision process (MDP). The MDP is characterized by a tuple $\mathcal{M}=\langle S, A, r, p, \gamma\rangle$, where $S$ is a finite set of states, $A$ is a finite set of actions, $r: S \times A \times S \rightarrow \mathbb{R}$ is the reward function, $p(s_{t+1} | s_{t}, a_{t})$ is the transition probability distribution, and $\gamma \in(0,1]$ is the discount factor. 
Given a state $s \in S$, a policy $\pi(a | s)$ is a probability distribution over the actions $a \in A$. At each time step $t$, the agent is in a state $s_t$, selects an action $a_t$ according to a policy $\pi(.| s_t)$, and executes the action. The agent then receives a new state $s_{t+1} \sim p(. | s_t, a_t)$ and  
a reward $r(s_t, a_t, s_{t+1})$ from the environment. 
The objective of the agent is to discover the optimal policy $\pi^{*}$ that maximizes the expected discounted return $G_{t}=\mathbb{E}_{\pi}[\sum_{k=0}^{\infty} \gamma^{k} r_{t+k}|S_{t}=s]$ for any state $s \in S$ and time step $t$.

The $Q$-function $q^\pi(s,a)$ under a policy $\pi$ is the expected discounted return of taking action $a$ in state $s$ and then following policy $\pi$. 
It is established that for every state $s \in S$ and action $a \in A$, every optimal policy $\pi^{*}$ satisfies the Bellman optimality equations (where $q^{*}=q^{\pi^{*}}$); $
q^{*}(s, a)=\sum_{s^{\prime} \in S} p(s^{\prime} | s, a)\left(r(s, a, s^{\prime})+\gamma \max _{a^{\prime} \in A} q^{*}(s^{\prime}, a^{\prime})\right)$.
It should be noted that if $q^{*}$ is known, selecting the action $a$ with the highest value of $q^{*}(s, a)$ always results in an optimal policy.

\textbf{Tabular $Q$-learning} \cite{watkins1992q} is a popular RL method, which estimates the optimal $Q$-function using the agent's experience. The estimated $Q$-value is denoted as $\tilde{q}(s, a)$. At each iteration, the agent observes the current state $s$ and chooses an action $a$ based on an exploratory policy. One commonly used exploratory policy is the $\epsilon$-greedy policy, which randomly selects an action with probability $\epsilon$, and chooses the action with the highest $\tilde{q}(s,a)$ value with probability $1-\epsilon$.
In this paper, as for $\epsilon$, we have employed an exponentially decaying version, where $\epsilon = \epsilon_{min} + (\epsilon_{max} - \epsilon_{min})\cdot\exp{(-decay \cdot episode counter)}$.

After the agent selects an action $a$ and transitions from state $s$ to $s'$, the resulting state $s^{\prime}$ and immediate reward $r(s,a,s^{\prime})$ are used to update the estimated $Q$-value of the current state-action pair $\tilde{q}(s,a)$. This is done using the $Q$-learning update rule; $\tilde{q}(s, a) \leftarrow \tilde{q}(s, a) + \alpha \cdot \left(r(s, a, s^{\prime})+\gamma \max _{a^{\prime}}
\tilde{q}(s^{\prime}, a^{\prime}) - \tilde{q}(s, a)\right),$
where $\alpha$ is the learning rate hyperparameter. If the resulting state $s^{\prime}$ is a terminal state, the update rule simplifies to $\tilde{q}(s, a) \leftarrow \tilde{q}(s, a) + \alpha \cdot \left(r(s, a, s^{\prime}) - \tilde{q}(s, a)\right)$.

The convergence of Tabular $Q$-learning to an optimal policy is guaranteed, provided that every state-action pair is visited infinitely often. As a learning method, this algorithm is classified as off-policy because it has the ability to learn from the experiences generated by any policy.

\subsection{NARS and ONA} \label{sectionNATRS}
\textbf{NARS} is an AI project that aims to create a general-purpose thinking machine. The underlying theory behind NARS is that intelligence is the ability for a system to adapt to its environment while working with insufficient knowledge and resources, as proposed by Wang \cite{wang1995non, wang2006rigid}.

NARS is a reasoning system that is based on the principles of Non-Axiomatic Logic (NAL). NAL is a formal logic that includes a formal language, Narsese, a set of formal inference rules, and semantics.
Conceptually, NAL is defined in a hierarchical manner, consisting of several layers, with each layer introducing new grammar and inference rules. This approach extends the logic and enhances its capability to express and infer, resulting in a more powerful reasoning system.
NAL allows for uncertainty and inconsistency in reasoning, making it more suitable for real-world applications where knowledge is often incomplete and uncertain \cite{wang2010non}.

NARS attempts to provide a normative model of general intelligence, rather than a descriptive model of human intelligence, although the latter can be seen as a special case of the former. Thus, while there may be some differences, the two types of models are similar in various aspects. 
The control component of NARS is mainly composed of a memory mechanism and an inference control mechanism \cite{wang2010non}.
The logic supports to reason on events coming from the agent's sensors in real-time, using an open-ended inference control process which does not terminate, whereby both forward (belief reasoning) and backward chaining (goal and question derivation) happen simultaneously.
The system draws conclusions from the available evidence in the premises, and then uses those conclusions to guide future reasoning and decision-making with a form of dynamic resource allocation, whereby only the most useful knowledge is kept in memory to satisfy a strictly bounded memory supply.

NARS represents knowledge as statements with attached truth and desire values, and uses inference rules to derive new knowledge from the premises, whereby truth functions are used to calculate conclusion evidence from the evidence summarized in the premises. In this system, to measure evidential support using relative measurements, a truth value is a pair of rational numbers in the range from 0 to 1. The first element of the truth value is frequency, and the second is confidence. Frequency is defined as the proportion of positive evidence among total evidence, that is, (positive evidence)/(total evidence).
Confidence indicates how sensitive the corresponding frequency is with respect to new evidence, as it is defined as the proportion of total evidence among total evidence plus a constant amount of new evidence, that is, (total evidence) / (total evidence + k) where k is a system parameter and in most discussions takes the default value of 1. Thus frequency can be seen as the degree of belief system has for the statement and confidence as the degree of belief for that estimation of frequency.
In this system, desire values have the same format as truth values, and indicate how much the system wants to achieve a statement (making it happen, essentially). The desire values of input goals can be assigned by the user to reflect their relative importance or take default values \cite{wang2010non}.

\textbf{ONA} is an implementation of NARS designed for real-world applications. Compared to OpenNARS, ONA includes firmer design decisions which make the software more effective for practical purposes. Additionally, ONA aims to make NARS more accessible to users and developers by providing a Python interface and a range of miscellaneous tools that can be used to build applications \cite{hammer2020opennars, hammer2021autonomy}.

Additionally, NARS and ONA use the same formal language called Narsese, which allows to express NAL statements. Narsese can represent beliefs, goals, and questions, and in ONA also the inference rules on the meta-level make use of it to be more easily editable. ONA also provides a simple standard-I/O interface, which can be used to interface with other programming languages and systems and to work with data sources which can stream in Narsese statements into the system \cite{hammer2020opennars, hammer2021autonomy}. In this publication, ONA was chosen as the implementation to compare with the tabular $Q$-learning algorithm.

\section{Setups and Environments} \label{section2}
\begin{figure}[t]
     \centering
     \begin{subfigure}{0.2\columnwidth}
         \centering
         \includegraphics[width=\columnwidth]{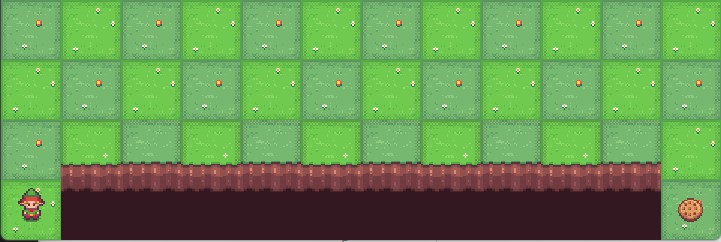}
         \caption{}
         \label{CliffWalking-v0}
     \end{subfigure}
     \begin{subfigure}{0.25\columnwidth}
         \centering
         \includegraphics[width=\columnwidth]{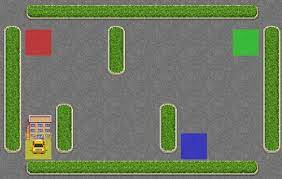}
         \caption{}
         \label{Taxi-v3}
     \end{subfigure}
     \begin{subfigure}{0.2\columnwidth}
         \centering
         \includegraphics[width=\columnwidth]{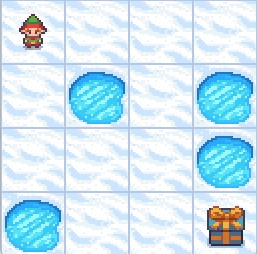}
         \caption{}
         \label{FrozenLake-v1 4x4}
     \end{subfigure}
    \begin{subfigure}{0.2\columnwidth}
         \centering
         \includegraphics[width=\columnwidth]{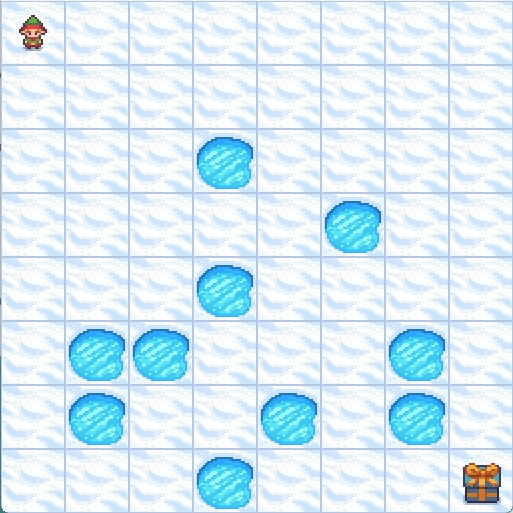}
         \caption{}
         \label{FrozenLake-v1 8x8}
     \end{subfigure}
     \begin{subfigure}{0.11\columnwidth}
         \centering
         \includegraphics[width=\columnwidth]{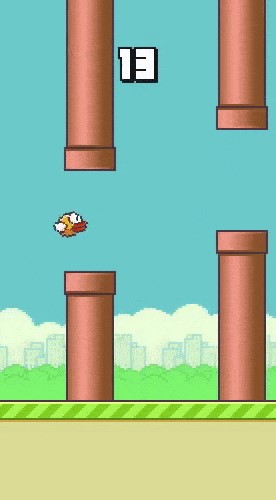}
         \caption{}
         \label{FlappyBird-v0}
     \end{subfigure}
        \caption{OpenAI gym environments used as experiment tasks; (a) CliffWalking-v0;(b) Taxi-v3;(c) FrozenLake-v1 4x4;(d) FrozenLake-v1 8x8;(e) FlappyBird-v0}
        \label{environments}
\end{figure}

Throughout the section, we describe how we implement both methods and compare their performance in different environments. 
Due to the stochastic nature of algorithms/environments and the dependency on hyperparameters, tabular $Q$-learning algorithm is notoriously difficult to evaluate. To comprehensively compare different algorithms, several environments, and network initialization seeds should be taken into account when tuning hyperparameters \cite{henderson2018deep}.
In this regard, to compare ONA with a tabular $Q$-learning \cite{watkins1992q} with exponentially decaying $\epsilon$ value, we conduct a grid search to tune the hyperparameters. For each combination of hyperparameters, we run the algorithm 10 times with different initialization and environment seeds. The configuration reported in the paper is the one that yielded the best performance on average among all tasks.
 In the case of $Q$-learning, we set $\alpha = 0.7$, $\gamma = 0.618$, $\epsilon_{max} = 1$, $\epsilon_{min} = 0.01$, decay $= 0.01$.
On the other hand, regarding ONA hyperparameters, specifically \textit{motorbabling}, we use the default value as used in ONA v0.9.1 \cite{hammer2020opennars}. However, \textit{babblingops} is changed due to the variety of the number of available actions in each of the environments. Also, we use \textit{setopname} to set allowed actions in ONA. The source code of our implementation is available online: \url{https://github.com/AliBeikmohammadi/OpenNARS-for-Applications/tree/master/misc/Python}

We primarily rely on the assumptions outlined in \cite{hammer2021autonomy}, unless explicitly stated otherwise.
To make the practical comparison possible, as for ONA, the events hold the same information as the corresponding states the $Q$-Learner receives in the simulated experiments, except for FlappyBird-v0. To be more specific, as mentioned in \cite{hammer2021autonomy}, when $s$ is an observation, it is interpreted by ONA as the event (s. $:|:$), and by the $Q$-Learner simply as the current state. Then both algorithms suggest an operation/action by exploitation or sometimes randomly. After feeding the action to the environment, we receive new observation, reward, and some info about reaching the goal. The reward for the $Q$-Learner is used without any change, while ONA receives an event (G. $:|:$) when the task is completely done. So there is no event if rewards are related to anything except finishing the task. This, of course, assumes that the goal does not change, as else the $Q$-table entries would have to be re-learned, meaning the learned behavior would often not apply anymore. For the purposes of this work, and for a fair comparison, the examples include a fixed objective.

We use challenging control tasks from OpenAI gym benchmark suite \cite{brockman2016openai} (Figure \ref{environments}). ONA and $Q$-Learning algorithms were developed for discrete tasks; hence we have to map FlappyBird-v0 observation space for each algorithm, which we describe below. Except for FlappyBird-v0, we used the original environments with no modifications to the environment or reward. In FlappyBird-v0, the observations are: ($O_1$) the horizontal distance to the next pipe, and ($O_2$) the difference between the player's $y$ position and the next hole's $y$ position. We have mapped this continuous observation space to a discrete space. Specifically, as for ONA, the event is "round(100x$O_1$)\_round(1000x$O_2$). $:|:$", which could be for instance "138\_-4. $:|:$". However, since for defining $Q$-table, the states should correspond to the specific row, we have to subtly change the mapping to "$|$round(100x$O_1$)$|$+$|$round(1000x$O_2$)$|$", which results "142", for our instance. However, one could find a better way to do this mapping.

Although \cite{brockman2016openai} describes all environments, we emphasize FrozenLake-v1's "is\_slippery" argument, allowing for a non-deterministic environment. This feature is interesting to observe how algorithms perform in such a scenario. When "is\_slippery" is True, the agent has a 1/3 probability of moving in the intended direction; otherwise, it moves in either perpendicular direction with an equal probability of 1/3. For instance, if the action is left and "is\_slippery" is True, then P(move left)=1/3, P(move up)=1/3, and P(move down)=1/3.
In the next section, we examine both algorithms' performances in detail on all these tasks.

\section{Results and Discussion} \label{section3}

\begin{figure}[t]
     \centering
     \begin{subfigure}{0.32\columnwidth}
         \centering
         \includegraphics[width=\columnwidth]{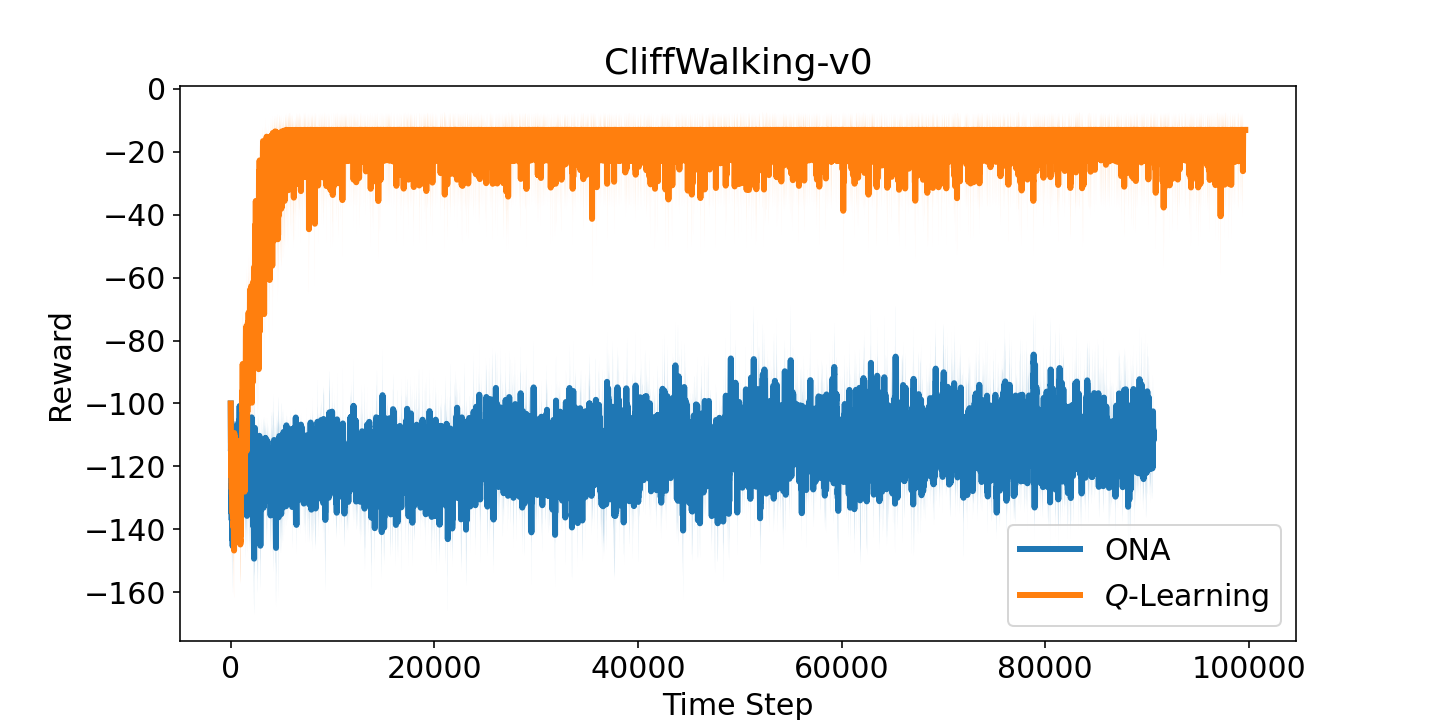}
         \caption{CliffWalking-v0}
         \label{Reward_vs_Time_Step_CliffWalking-v0}
     \end{subfigure}
     \begin{subfigure}{0.32\columnwidth}
         \centering
         \includegraphics[width=\columnwidth]{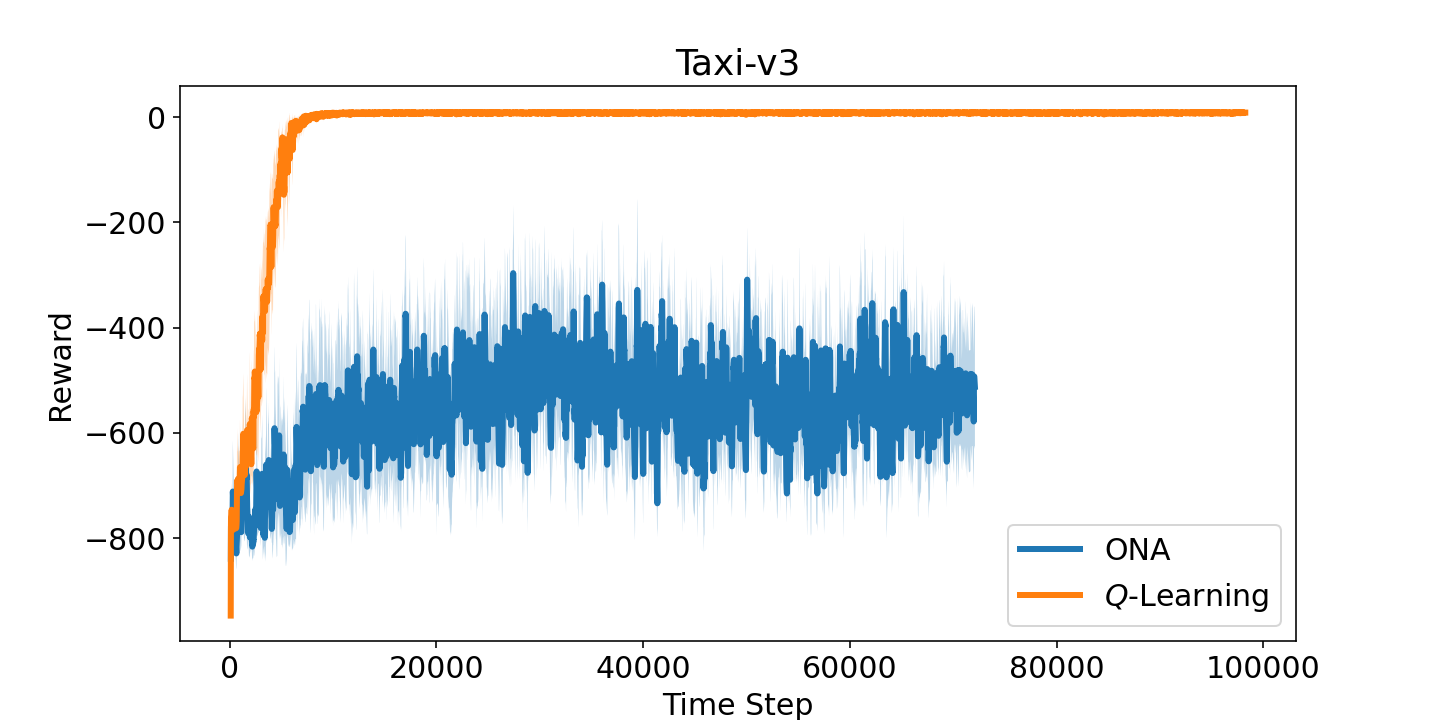}
         \caption{Taxi-v3}
         \label{Reward_vs_Time_Step_Taxi-v3}
     \end{subfigure}
     \begin{subfigure}{0.32\columnwidth}
         \centering
         \includegraphics[width=\columnwidth]{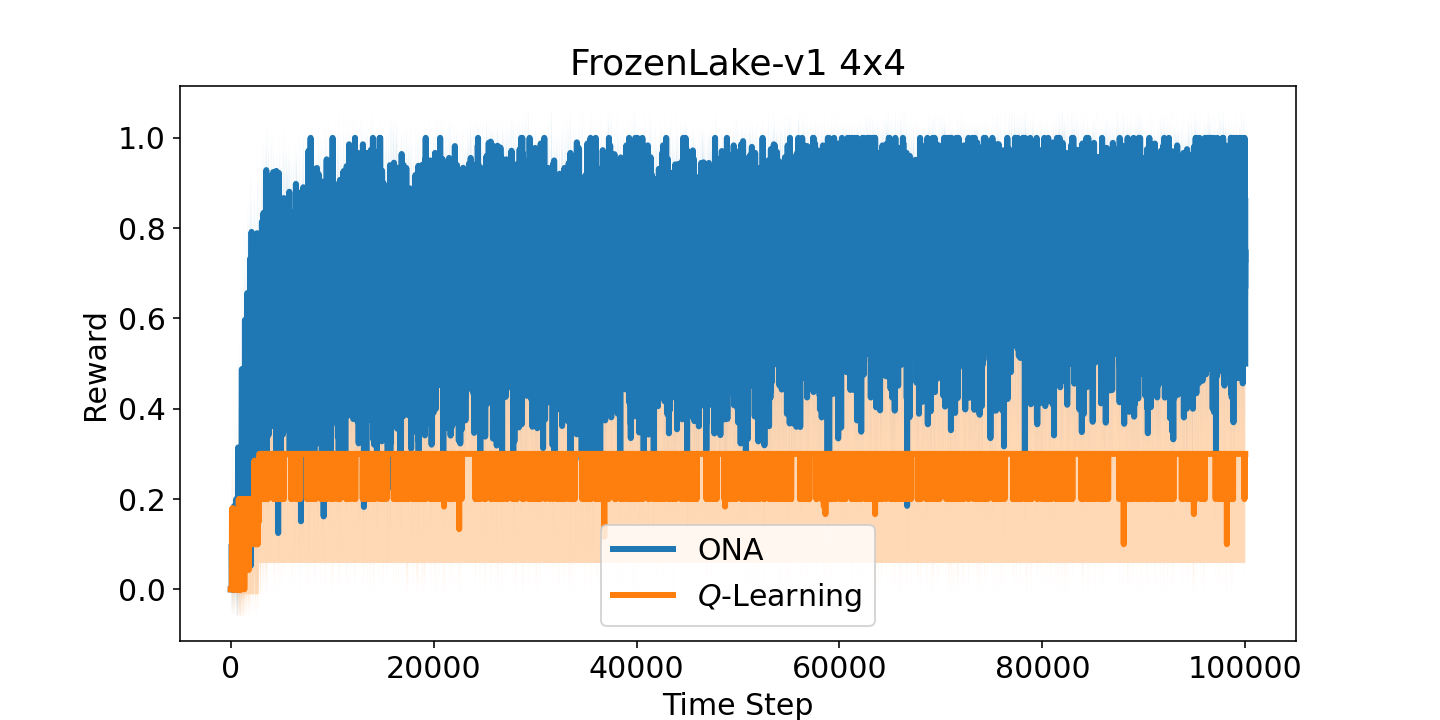}
         \caption{FrozenLake-v1 4x4}
         \label{Reward_vs_Time_Step_FrozenLake-v1 4x4}
     \end{subfigure}
     \begin{subfigure}{0.32\columnwidth}
         \centering
         \includegraphics[width=\columnwidth]{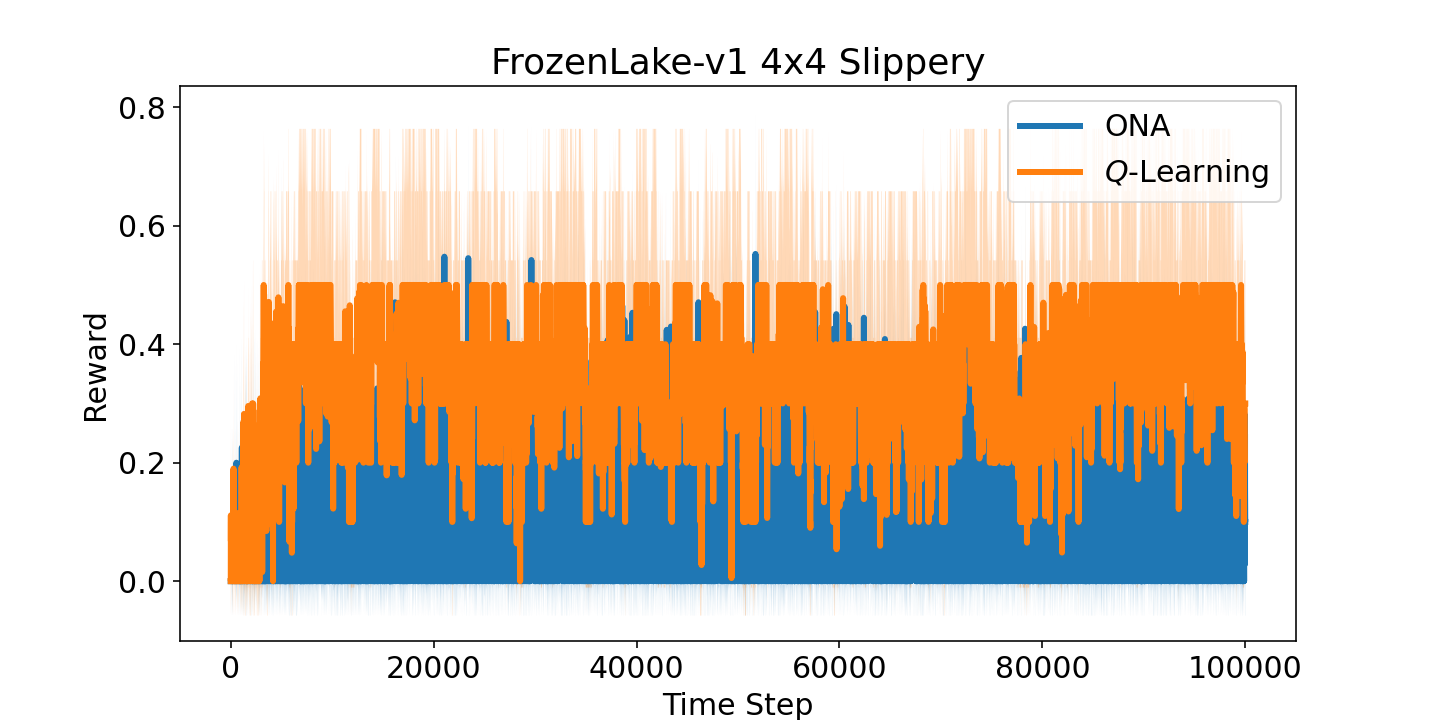}
         \caption{\scriptsize FrozenLake-v1 4x4 Slippery}
         \label{Reward_vs_Time_Step_FrozenLake-v1_4x4_Slippery}
     \end{subfigure}
    \begin{subfigure}{0.32\columnwidth}
         \centering
         \includegraphics[width=\columnwidth]{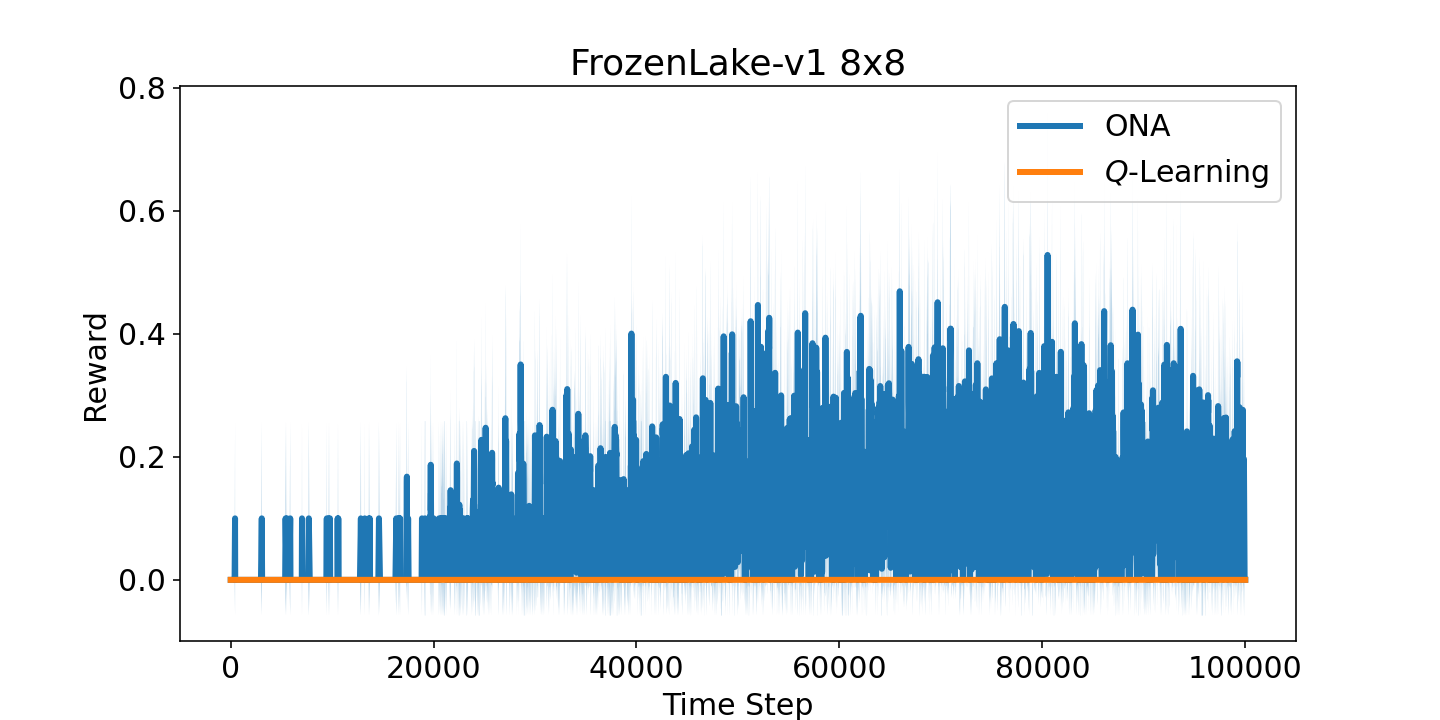}
         \caption{FrozenLake-v1 8x8}
         \label{Reward_vs_Time_Step_FrozenLake-v1 8x8}
     \end{subfigure}
    \begin{subfigure}{0.32\columnwidth}
         \centering
         \includegraphics[width=\columnwidth]{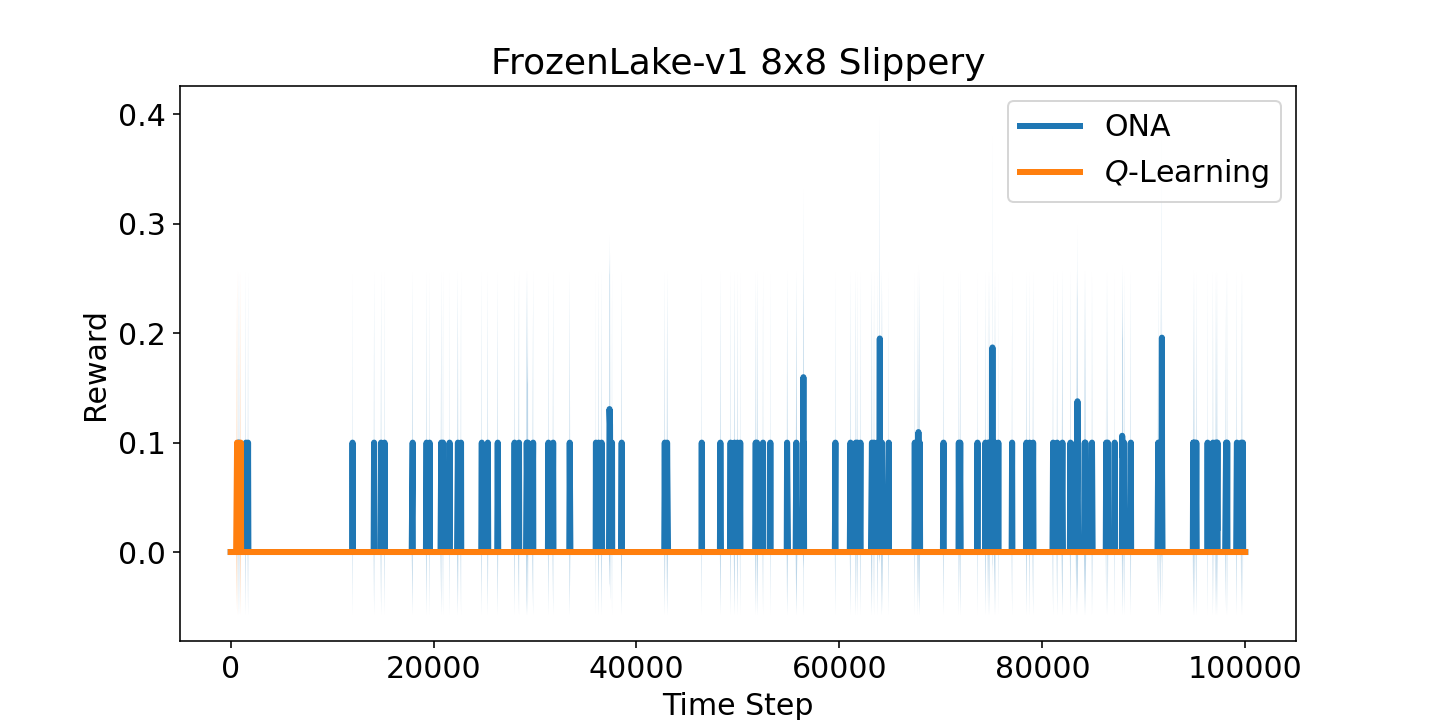}
          \caption{\scriptsize FrozenLake-v1 8x8 Slippery}
         \label{Reward_vs_Time_Step_FrozenLake-v1_8x8_Slippery}
     \end{subfigure}
     \begin{subfigure}{0.32\columnwidth}
         \centering
         \includegraphics[width=\columnwidth]{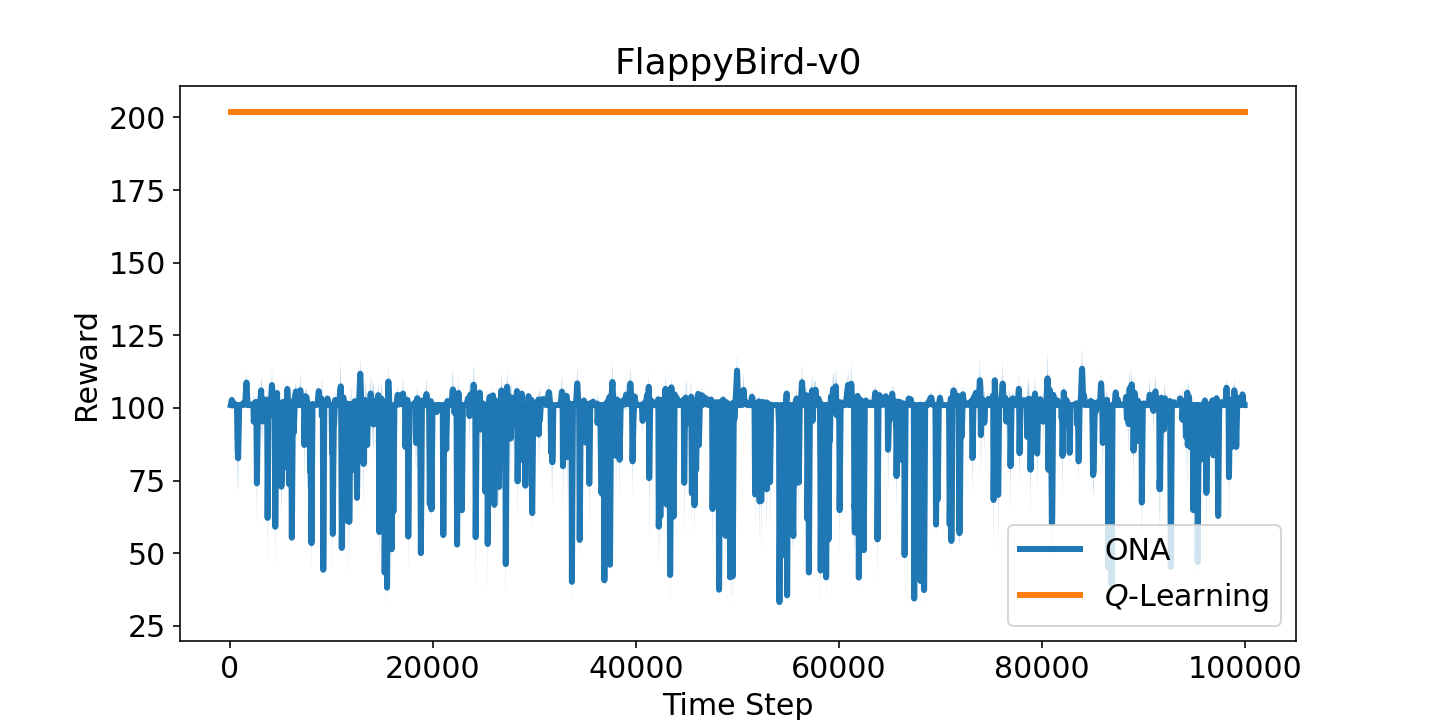}
         \caption{FlappyBird-v0}
         \label{Reward_vs_Time_Step_FlappyBird-v0}
     \end{subfigure}
        \caption{Reward vs. Time steps. The reward is measured at time steps where the episode ends (by reaching the goal, truncating the episode length, falling into the hole, falling from the cliff, hitting the pipe.)}
        \label{Reward_vs_Time Step}
\end{figure}

\begin{figure}[htp!]
     \centering
     \begin{subfigure}{0.32\columnwidth}
         \centering
         \includegraphics[width=\columnwidth]{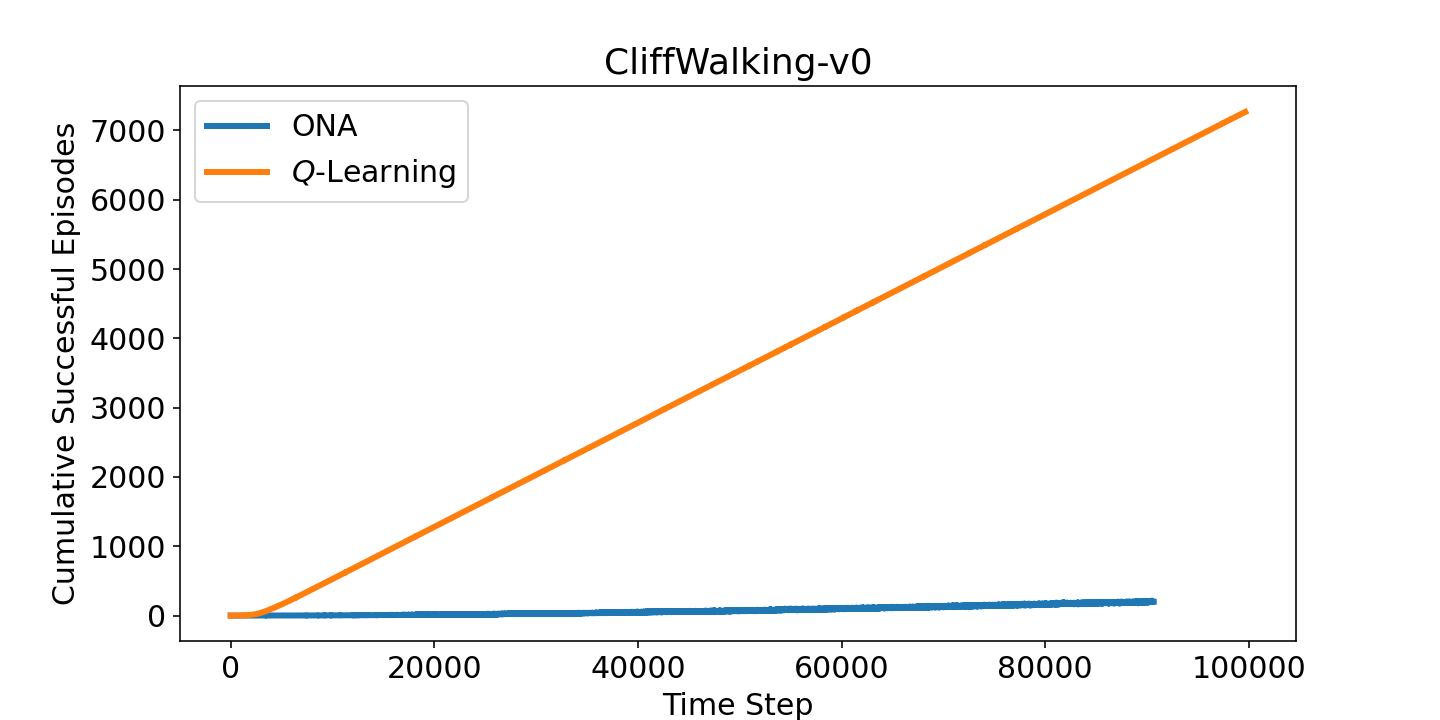}
         \caption{CliffWalking-v0}
         \label{Cumulative_Successful_Episodes_vs_Time_Step_CliffWalking-v0}
     \end{subfigure}
     \begin{subfigure}{0.32\columnwidth}
         \centering
         \includegraphics[width=\columnwidth]{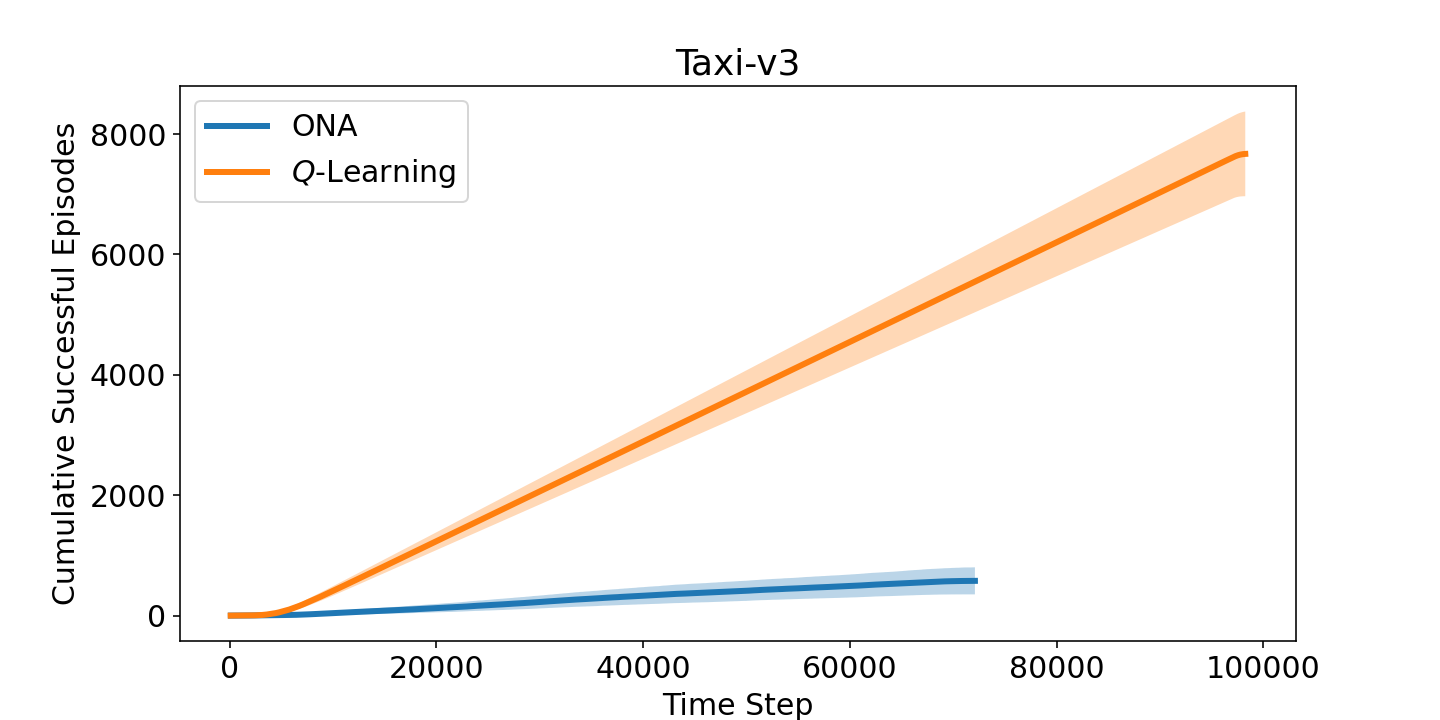}
         \caption{Taxi-v3}
         \label{Cumulative_Successful_Episodes_vs_Time_Step_Taxi-v3}
     \end{subfigure}
     \begin{subfigure}{0.32\columnwidth}
         \centering
         \includegraphics[width=\columnwidth]{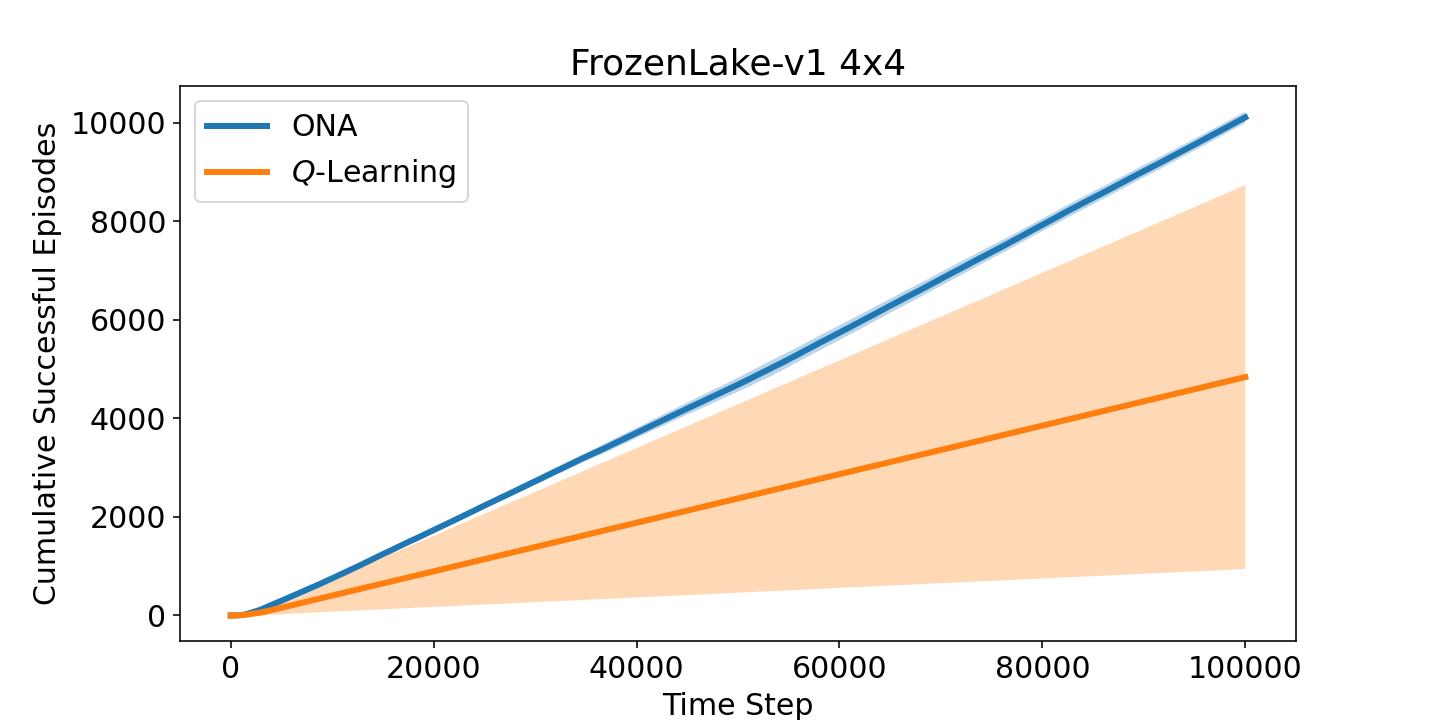}
         \caption{FrozenLake-v1 4x4}
         \label{Cumulative_Successful_Episodes_vs_Time_Step_FrozenLake-v1 4x4}
     \end{subfigure}
     \begin{subfigure}{0.32\columnwidth}
         \centering
         \includegraphics[width=\columnwidth]{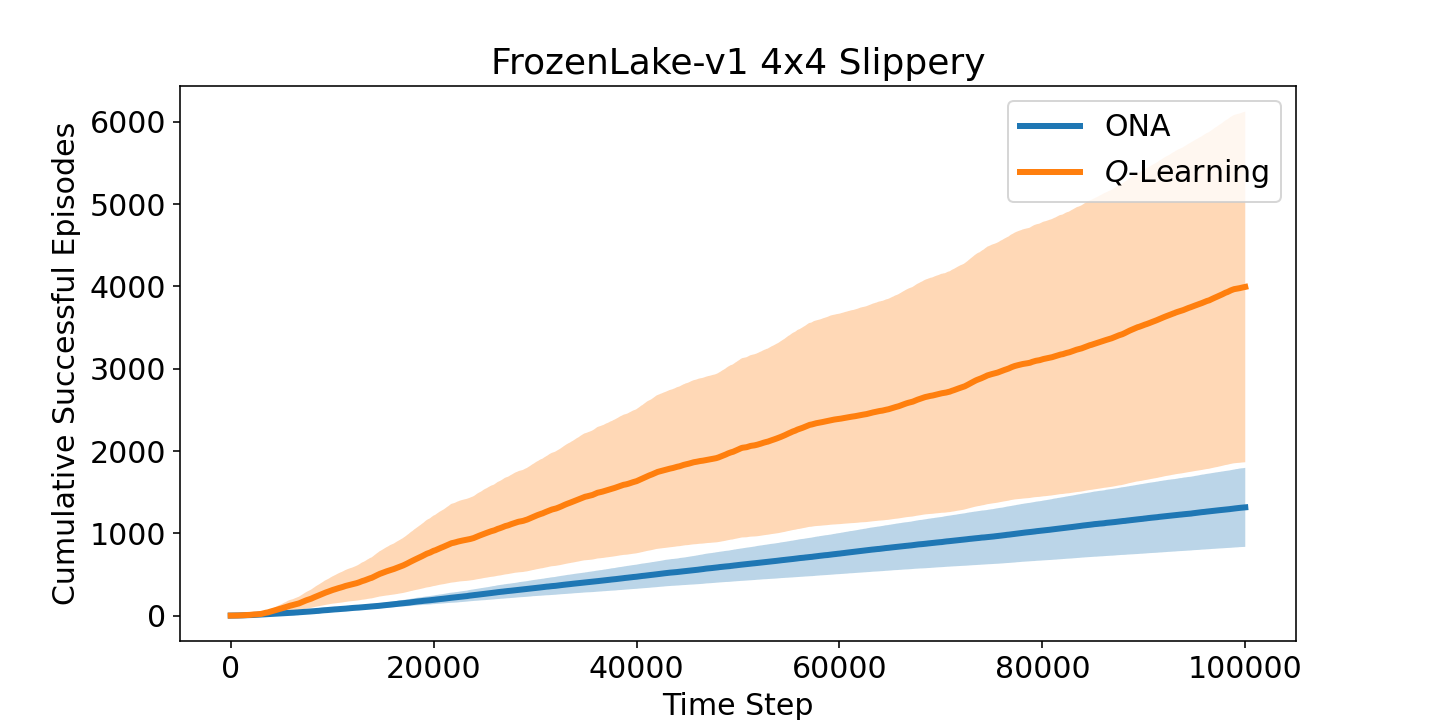}
         \caption{\scriptsize FrozenLake-v1 4x4 Slippery}
         \label{Cumulative_Successful_Episodes_vs_Time_Step_FrozenLake-v1_4x4_Slippery}
     \end{subfigure}
    \begin{subfigure}{0.32\columnwidth}
         \centering
         \includegraphics[width=\columnwidth]{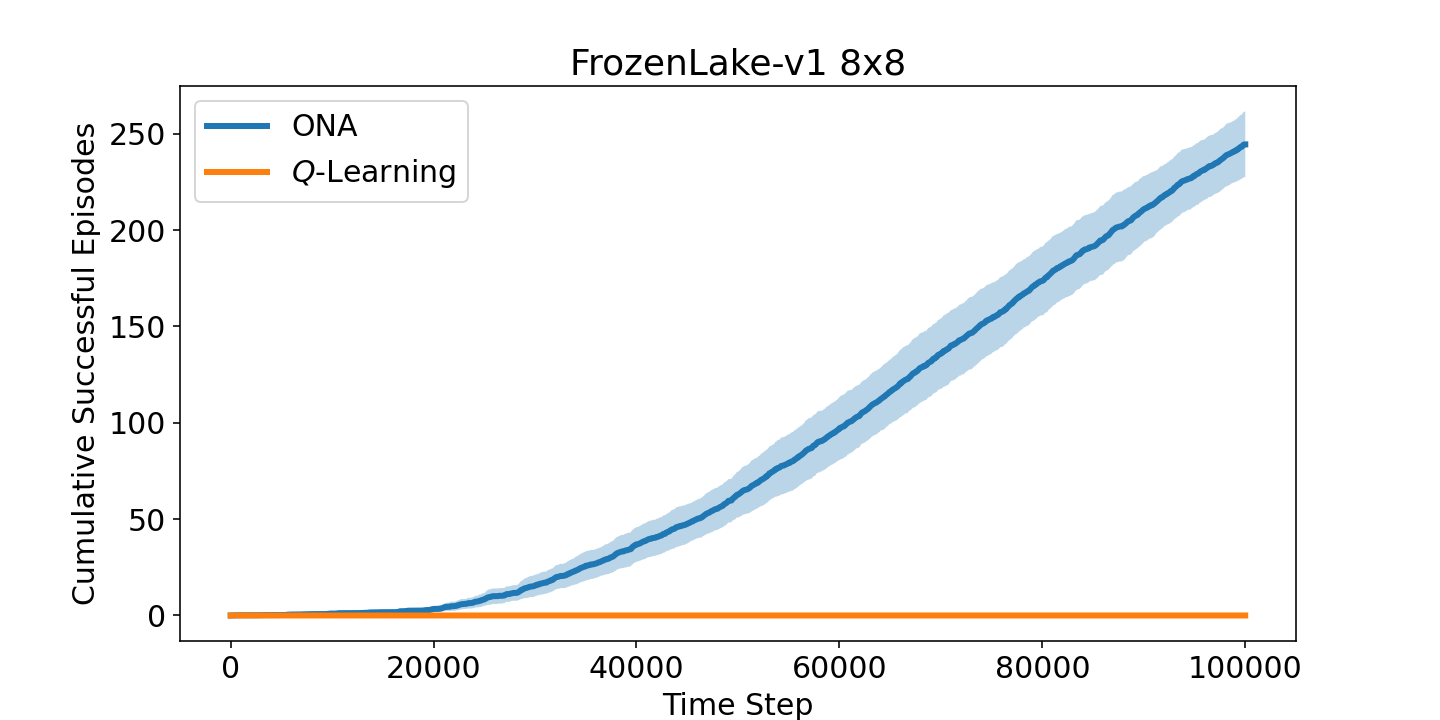}
         \caption{FrozenLake-v1 8x8}
         \label{Cumulative_Successful_Episodes_vs_Time_Step_FrozenLake-v1 8x8}
     \end{subfigure}
    \begin{subfigure}{0.32\columnwidth}
         \centering
         \includegraphics[width=\columnwidth]{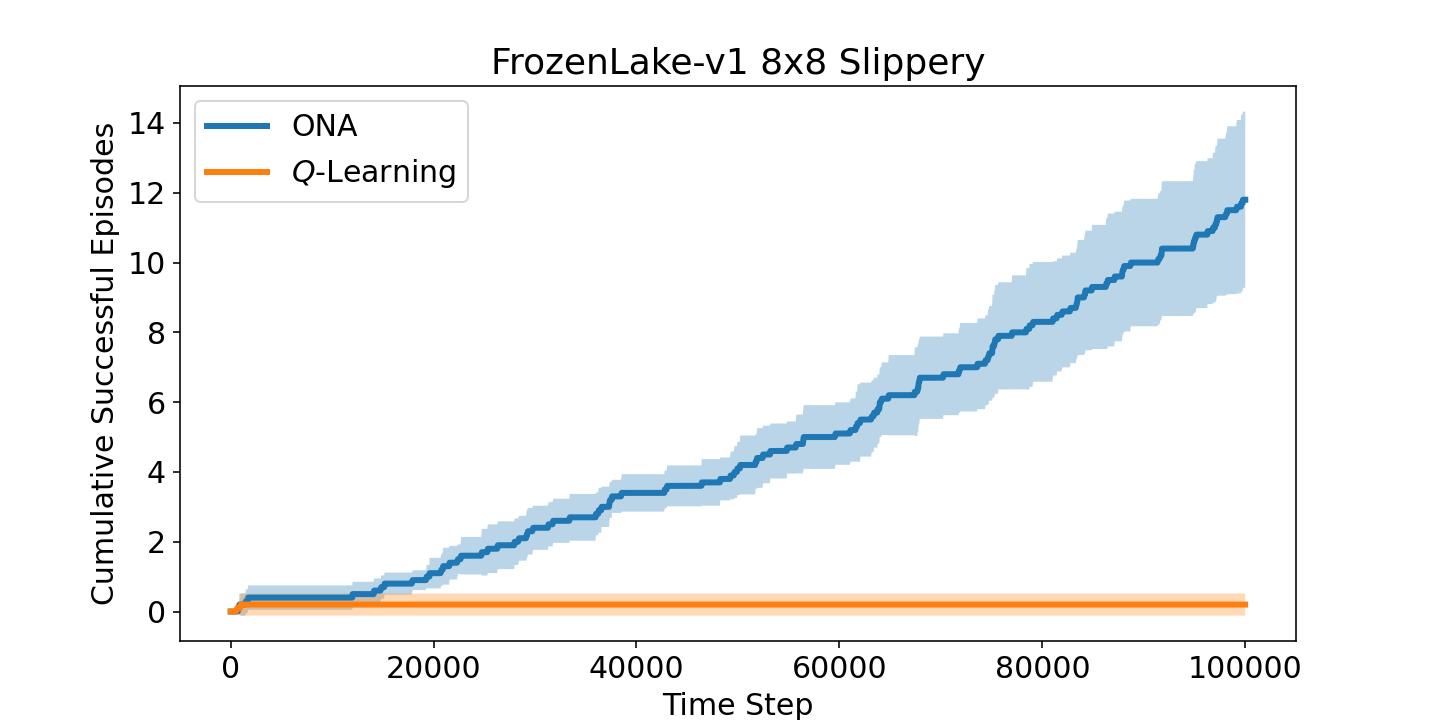}
         \caption{\scriptsize FrozenLake-v1 8x8 Slippery}
         \label{Cumulative_Successful_Episodes_vs_Time_Step_FrozenLake-v1_8x8_Slippery}
     \end{subfigure}
        \caption{Cumulative Successful Episodes vs. Time steps.}
        \label{Cumulative_Successful_Episodes_vs_Time Step}
\end{figure}

Two criteria, including reward, and cumulative successful episodes, are monitored, as shown in Figures \ref{Reward_vs_Time Step}, and \ref{Cumulative_Successful_Episodes_vs_Time Step}.
Both techniques are run 10 times in each experiment, and the behavior of metrics is kept track of for each time step across 100000 iterations. The solid curves show average training performance, while the shaded region indicates the standard deviation of that specific metric over 10 trials. This provides an idea of the algorithm's robustness. A high gained metric with a low variance is considered more reliable than achieving the same performance with a high variance. So, the standard deviation gives a complete picture of the algorithm's performance.

As can be seen from Figures \ref{Reward_vs_Time Step} and \ref{Cumulative_Successful_Episodes_vs_Time Step}, the results of two algorithms are very dependent on the task and one cannot be considered superior for all environments. Specifically, the $Q$-Learning algorithm has performed better on CliffWalking-v0, Taxi-v3, and FlappyBird-v0 environments. But ONA is more promising on environments based on FrozenLake-v1. Moreover, Figure \ref{Cumulative_Successful_Episodes_vs_Time Step} illustrates the noteworthy observation that ONA exhibits greater reliability.


An interesting observation is the good ability of ONA to solve non-deterministic problems, where it is able to solve the slippery-enable problems as shown in Figures \ref{Reward_vs_Time_Step_FrozenLake-v1_4x4_Slippery},  \ref{Reward_vs_Time_Step_FrozenLake-v1_8x8_Slippery}, \ref{Cumulative_Successful_Episodes_vs_Time_Step_FrozenLake-v1_4x4_Slippery}, and \ref{Cumulative_Successful_Episodes_vs_Time_Step_FrozenLake-v1_8x8_Slippery}, while $Q$-Learning has not shown reliable success in solving these problems.
It may be possible to draw conclusions from $Q$-Learning by adjusting its hyperparameters. However, it should be noted that any time-dependent hyperparameters are specific to the environment and should be avoided when evaluating generality. Additionally, as $\epsilon$ decreases over time, the $Q$-Learner will take longer to adapt its policy to new circumstances. 
In contrast, it is evident that ONA offers greater reliability due to having fewer hyperparameters. Unlike $Q$-Learning, ONA does not require specific reductions in learning or exploration rates to function well on a given task, and therefore needs less parameter tuning. For instance, ONA does not rely on learning rate decay. Instead, the extent to which new evidence alters an existing belief is dependent solely on the degree of evidence that already supports it, which automatically renders high-confidence beliefs more stable. This results in a more consistent learning behavior for ONA.

\begin{figure}[t]
     \centering
     \begin{subfigure}{0.32\columnwidth}
         \centering
         \includegraphics[width=\columnwidth]{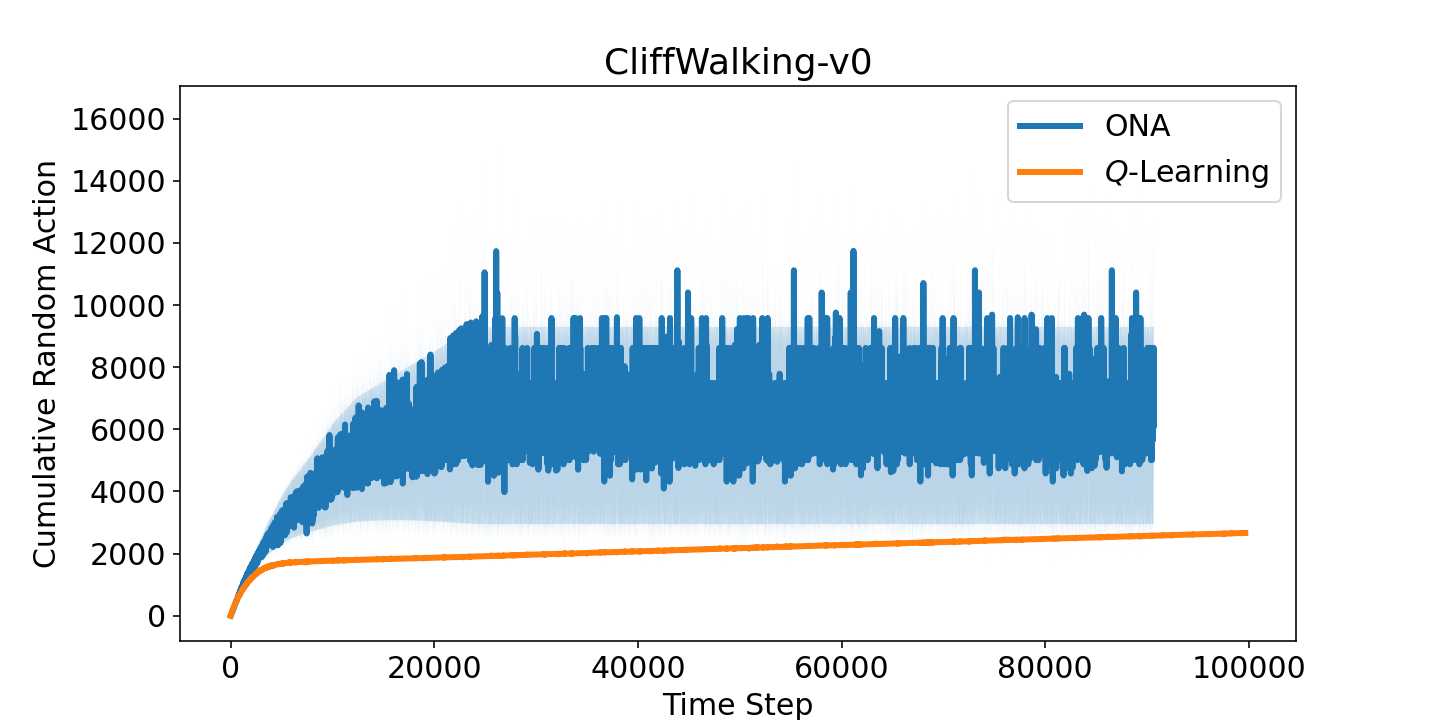}
         \caption{CliffWalking-v0}
         \label{Cumulative_Random_Action_vs_Time_Step_CliffWalking-v0}
     \end{subfigure}
     \begin{subfigure}{0.32\columnwidth}
         \centering
         \includegraphics[width=\columnwidth]{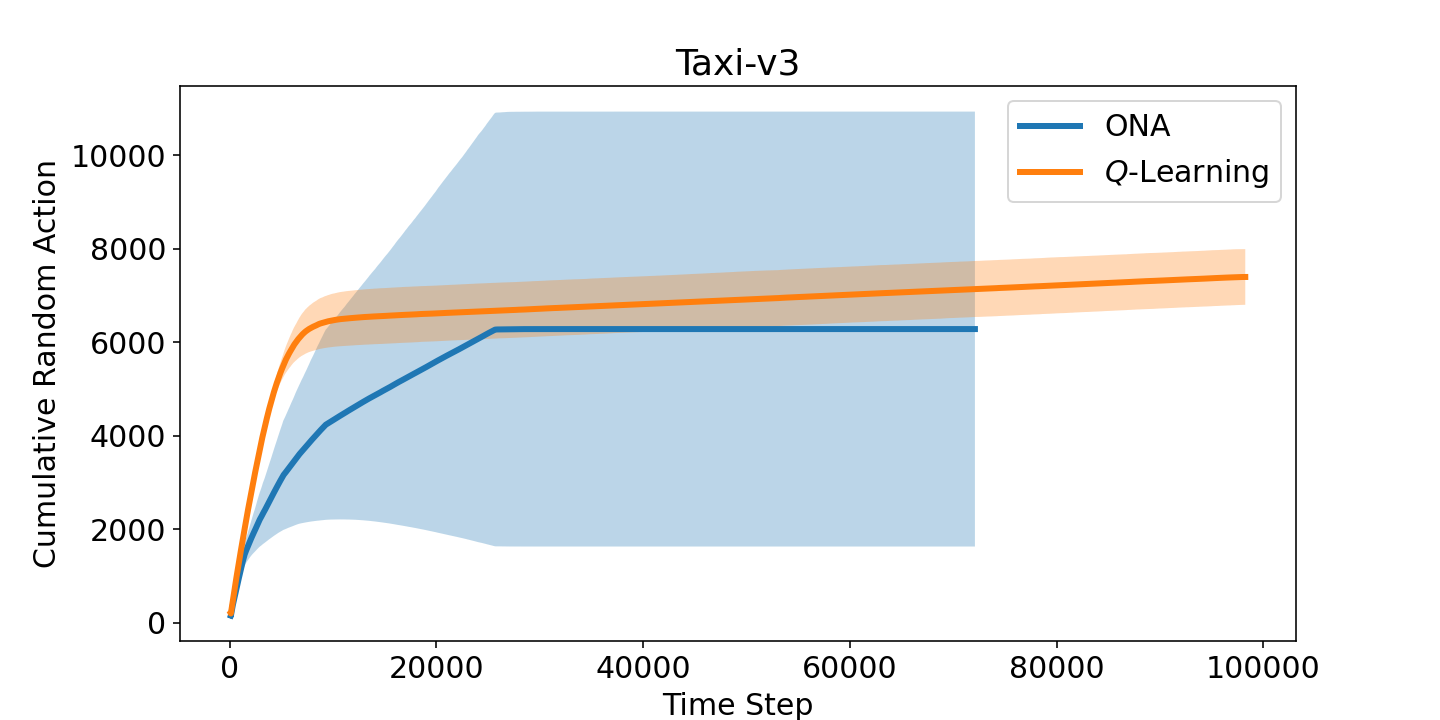}
         \caption{Taxi-v3}
         \label{Cumulative_Random_Action_vs_Time_Step_Taxi-v3}
     \end{subfigure}
     \begin{subfigure}{0.32\columnwidth}
         \centering
         \includegraphics[width=\columnwidth]{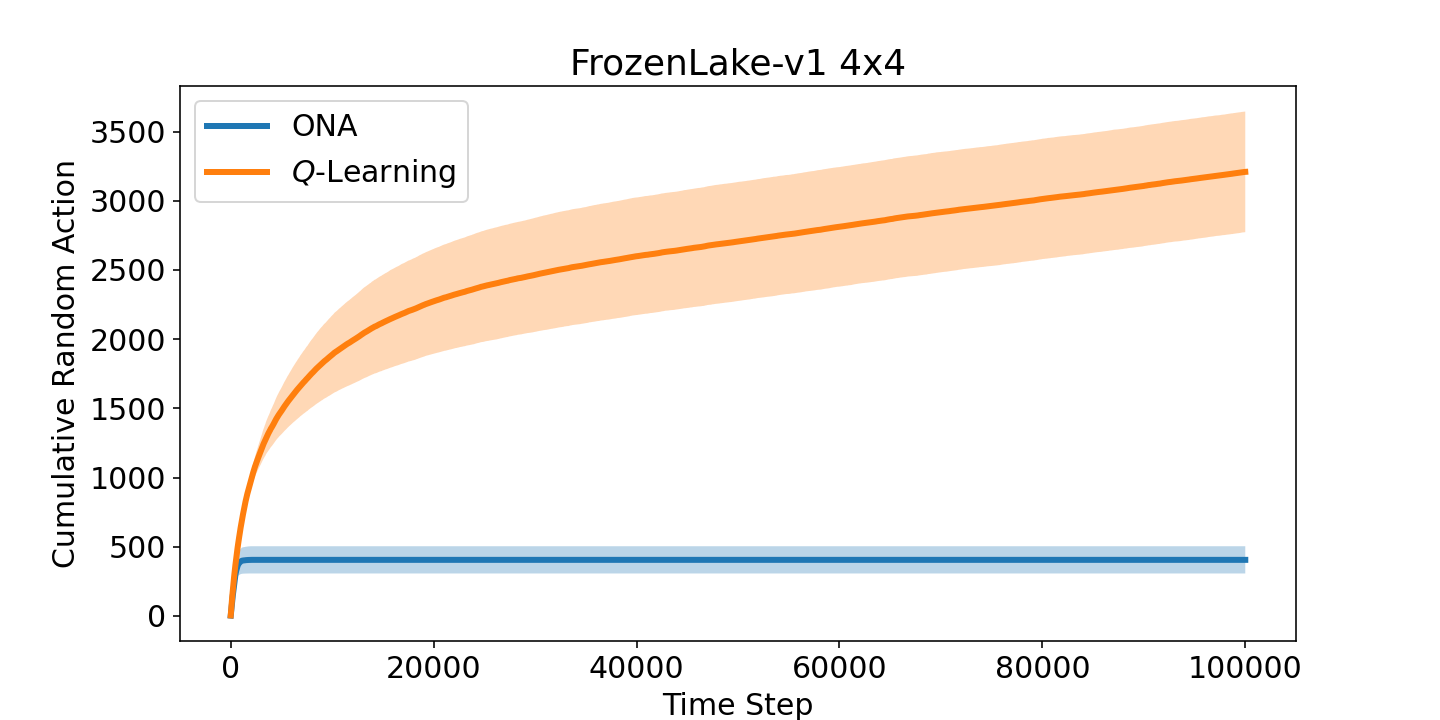}
         \caption{FrozenLake-v1 4x4}
         \label{Cumulative_Random_Action_vs_Time_Step_FrozenLake-v1 4x4}
     \end{subfigure}
     \begin{subfigure}{0.32\columnwidth}
         \centering
         \includegraphics[width=\columnwidth]{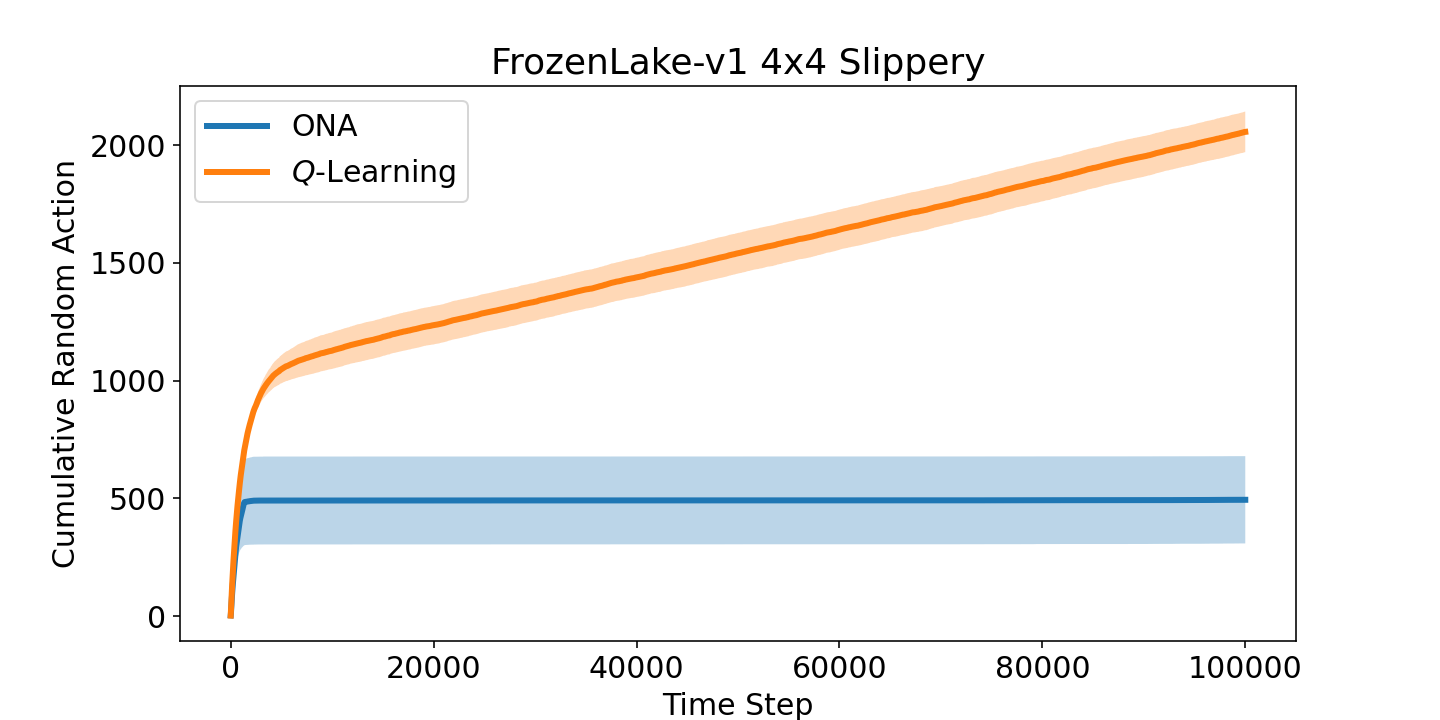}
         \caption{\scriptsize FrozenLake-v1 4x4 Slippery}
         \label{Cumulative_Random_Action_vs_Time_Step_FrozenLake-v1_4x4_Slippery}
     \end{subfigure}
    \begin{subfigure}{0.32\columnwidth}
         \centering
         \includegraphics[width=\columnwidth]{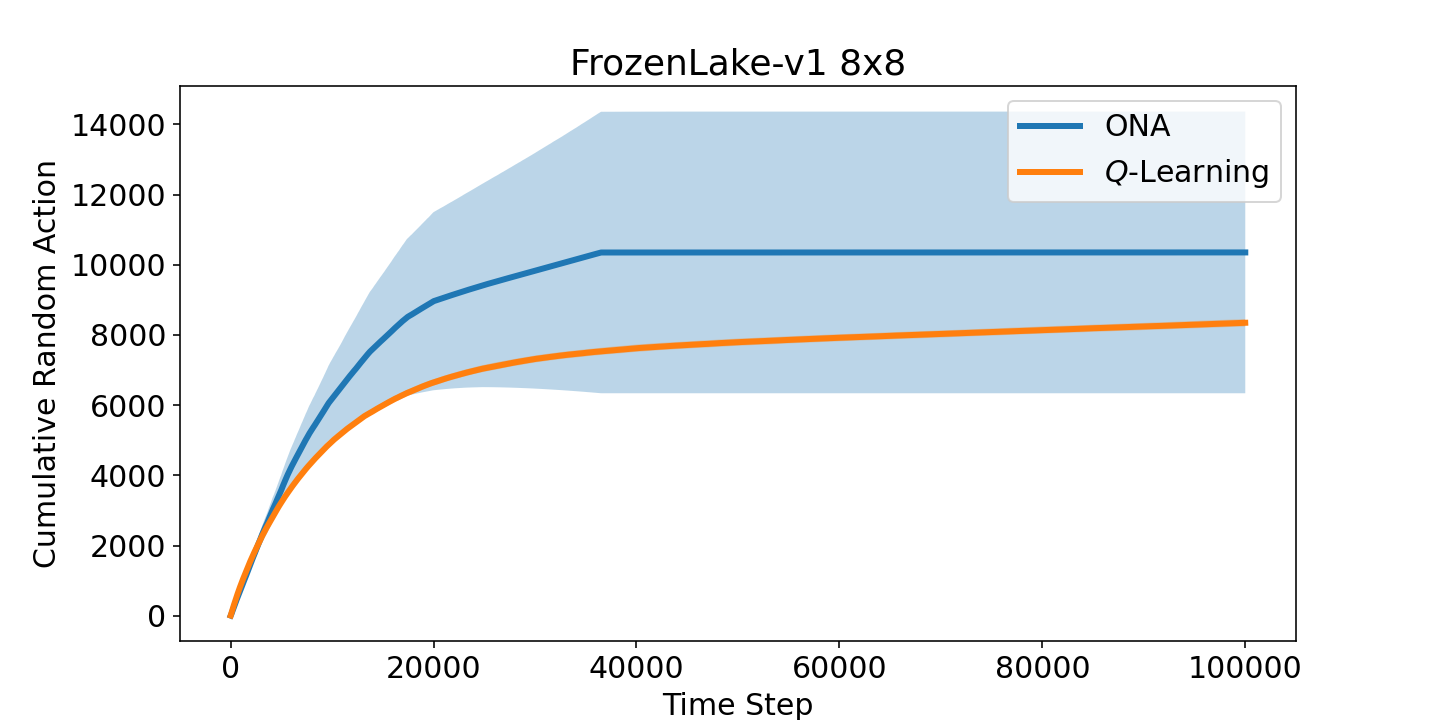}
         \caption{FrozenLake-v1 8x8}
         \label{Cumulative_Random_Action_vs_Time_Step_FrozenLake-v1 8x8}
     \end{subfigure}
    \begin{subfigure}{0.32\columnwidth}
         \centering
         \includegraphics[width=\columnwidth]{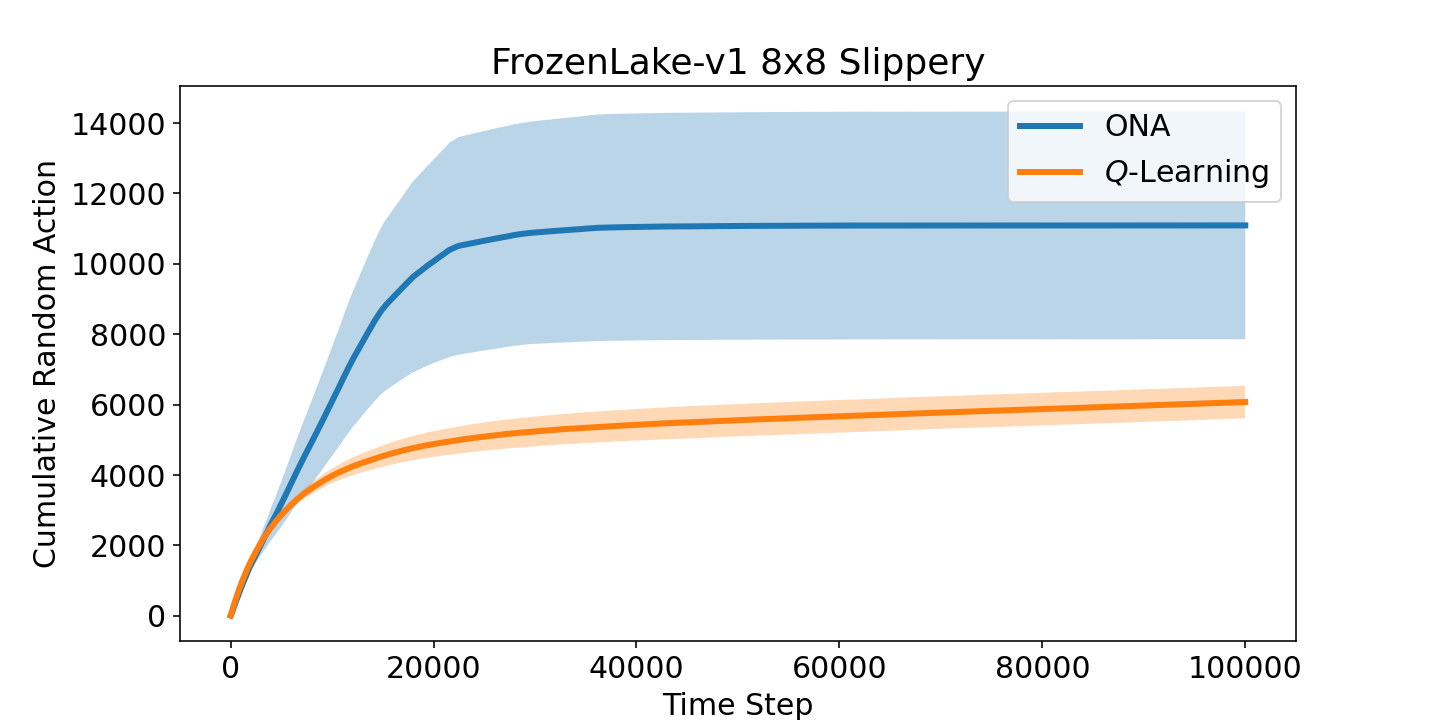}
         \caption{\scriptsize FrozenLake-v1 8x8 Slippery}
         \label{Cumulative_Random_Action_vs_Time_Step_FrozenLake-v1_8x8_Slippery}
     \end{subfigure}
     \begin{subfigure}{0.32\columnwidth}
         \centering
         \includegraphics[width=\columnwidth]{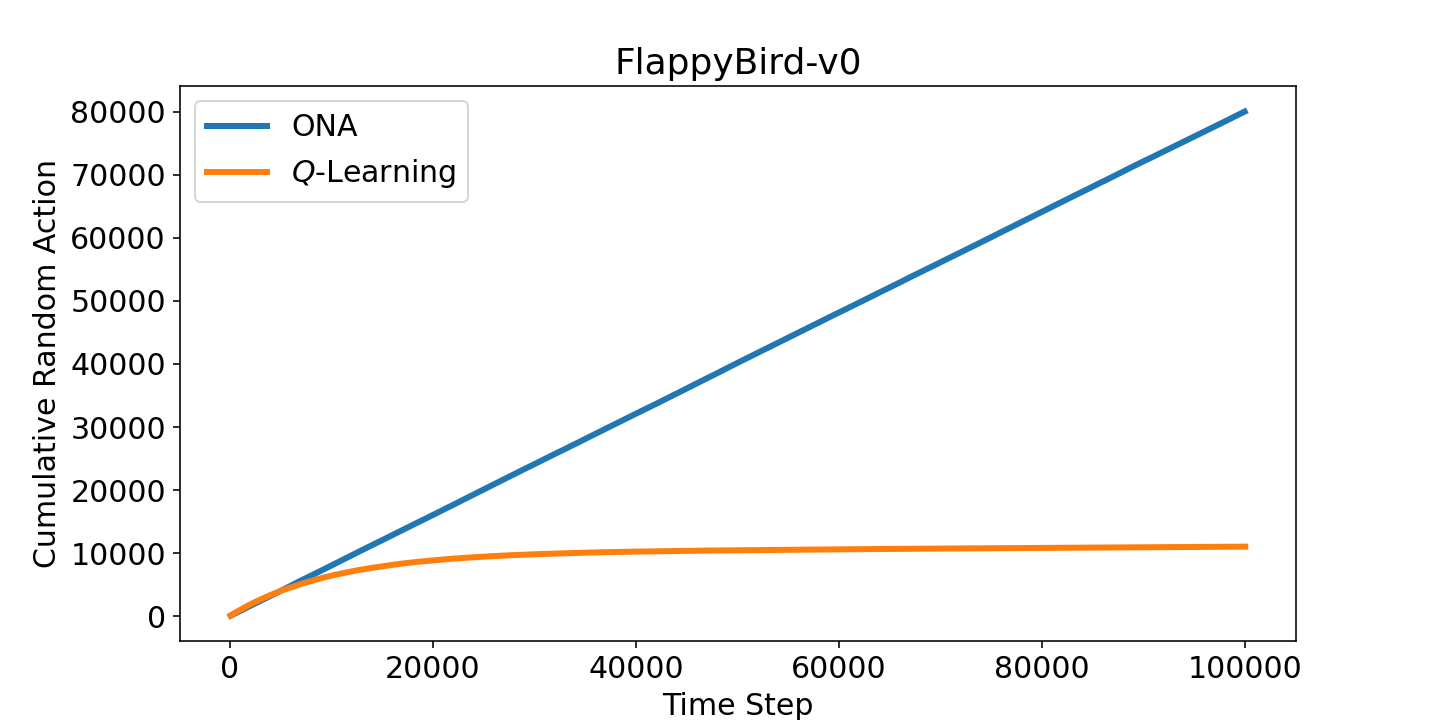}
         \caption{FlappyBird-v0}
         \label{Cumulative_Random_Action_vs_Time_Step_FlappyBird-v0}
     \end{subfigure}
        \caption{Cumulative Random Action vs. Time steps.}
        \label{Cumulative_Random_Action_vs_Time Step}
\end{figure}
\begin{figure}[t]
     \centering
     \begin{subfigure}{0.32\columnwidth}
         \centering
         \includegraphics[width=\columnwidth]{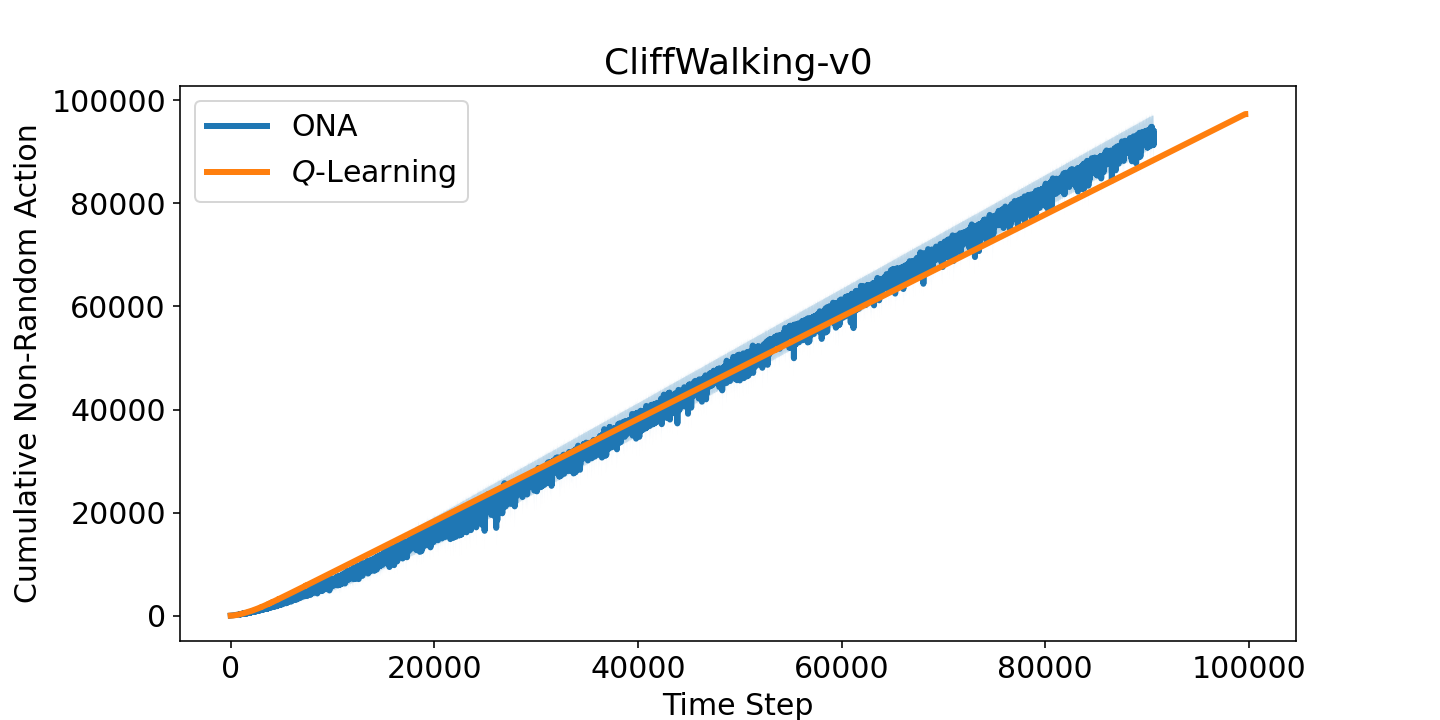}
         \caption{CliffWalking-v0}
         \label{Cumulative_Non-Random_Action_vs_Time_Step_CliffWalking-v0}
     \end{subfigure}
     \begin{subfigure}{0.32\columnwidth}
         \centering
         \includegraphics[width=\columnwidth]{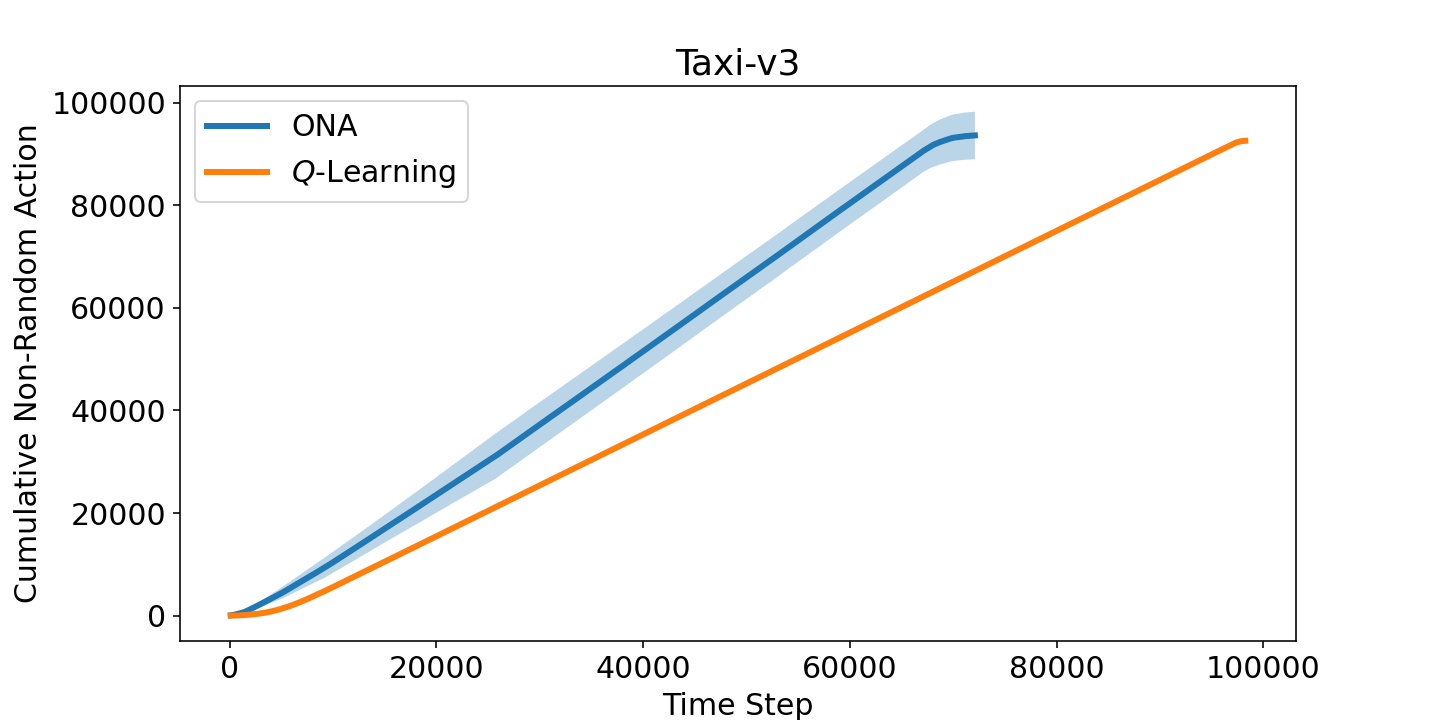}
         \caption{Taxi-v3}
         \label{Cumulative_Non-Random_Action_vs_Time_Step_Taxi-v3}
     \end{subfigure}
     \begin{subfigure}{0.32\columnwidth}
         \centering
         \includegraphics[width=\columnwidth]{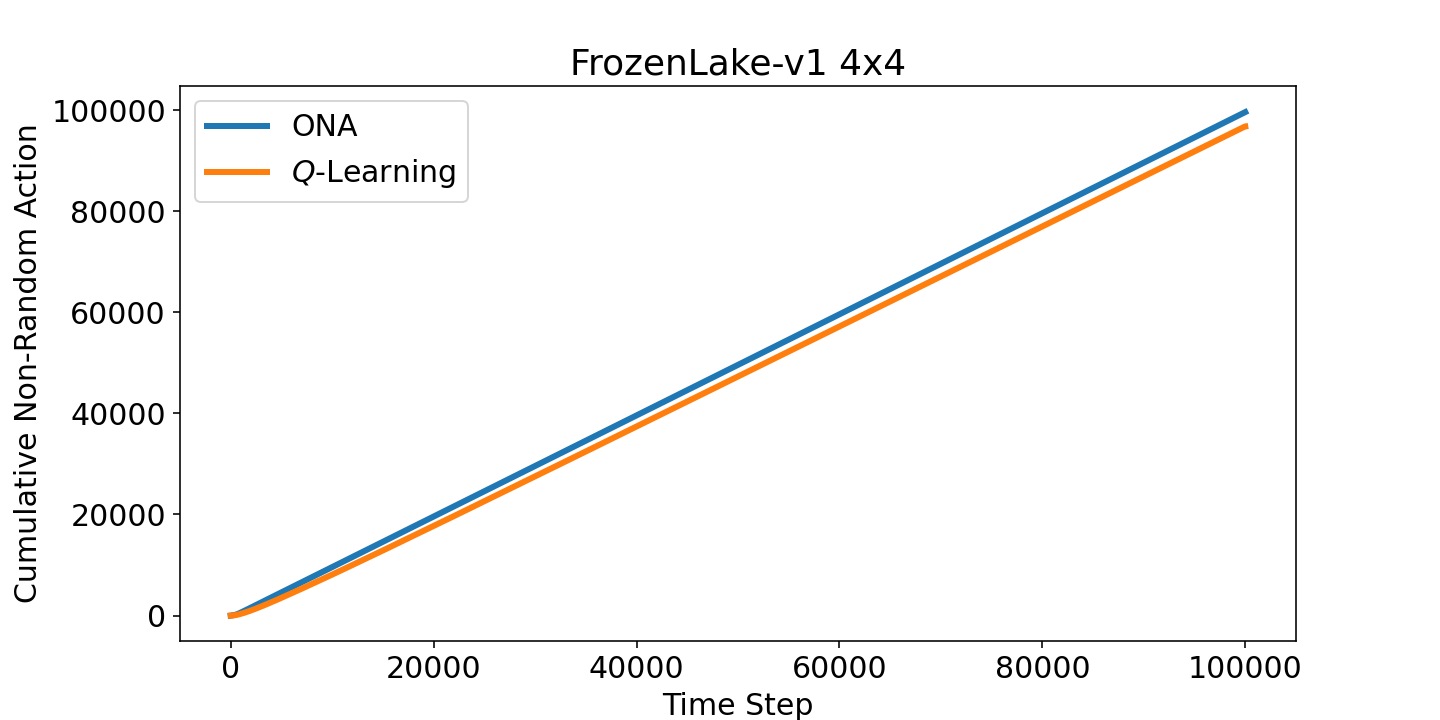}
         \caption{FrozenLake-v1 4x4}
         \label{Cumulative_Non-Random_Action_vs_Time_Step_FrozenLake-v1 4x4}
     \end{subfigure}
     \begin{subfigure}{0.32\columnwidth}
         \centering
         \includegraphics[width=\columnwidth]{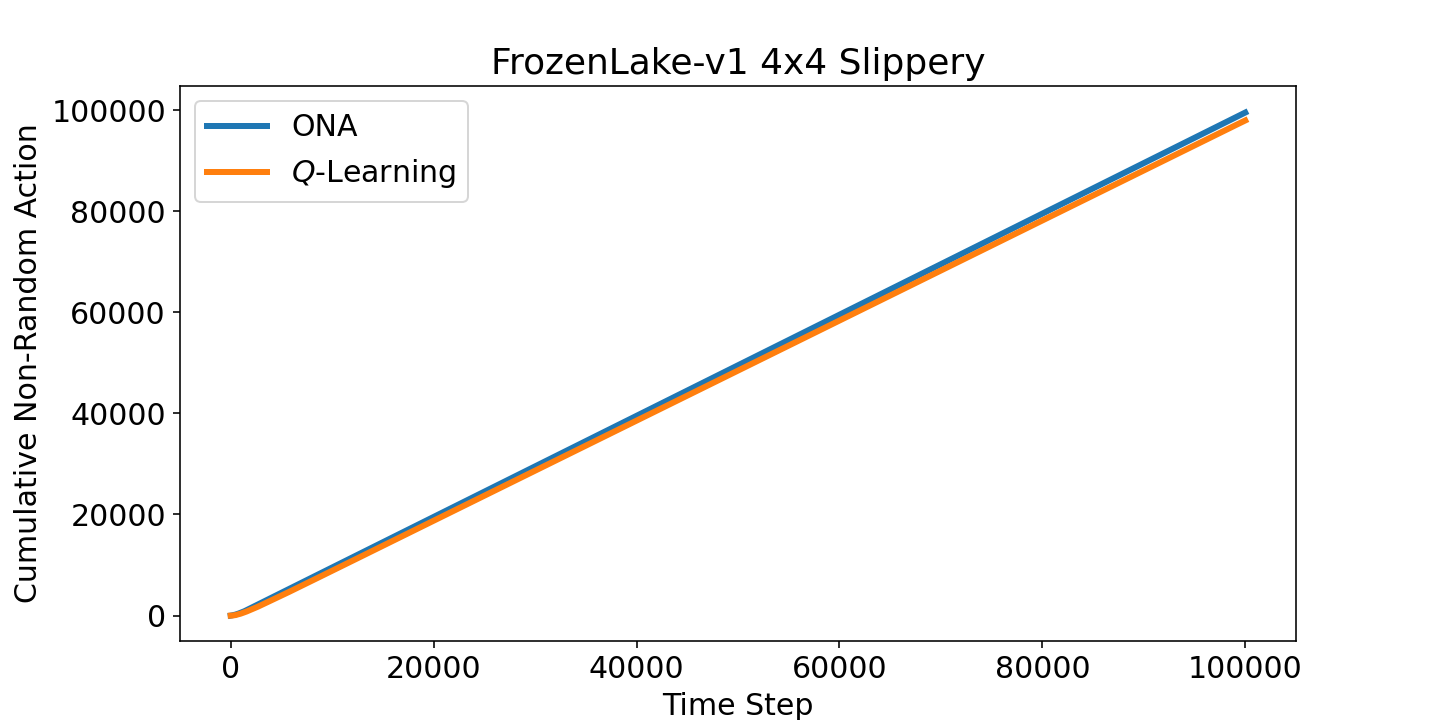}
         \caption{\scriptsize FrozenLake-v1 4x4 Slippery}
         \label{Cumulative_Non-Random_Action_vs_Time_Step_FrozenLake-v1_4x4_Slippery}
     \end{subfigure}
    \begin{subfigure}{0.32\columnwidth}
         \centering
         \includegraphics[width=\columnwidth]{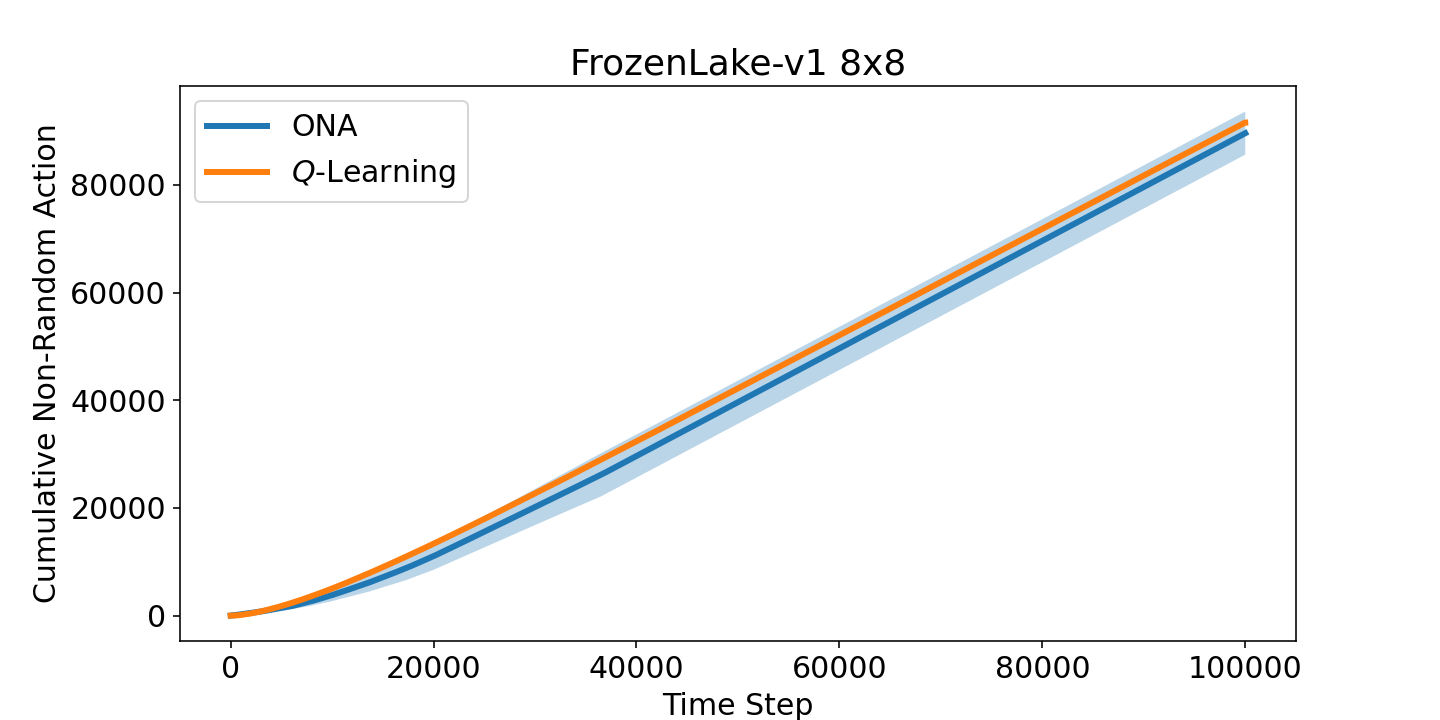}
         \caption{FrozenLake-v1 8x8}
         \label{Cumulative_Non-Random_Action_vs_Time_Step_FrozenLake-v1 8x8}
     \end{subfigure}
    \begin{subfigure}{0.32\columnwidth}
         \centering
         \includegraphics[width=\columnwidth]{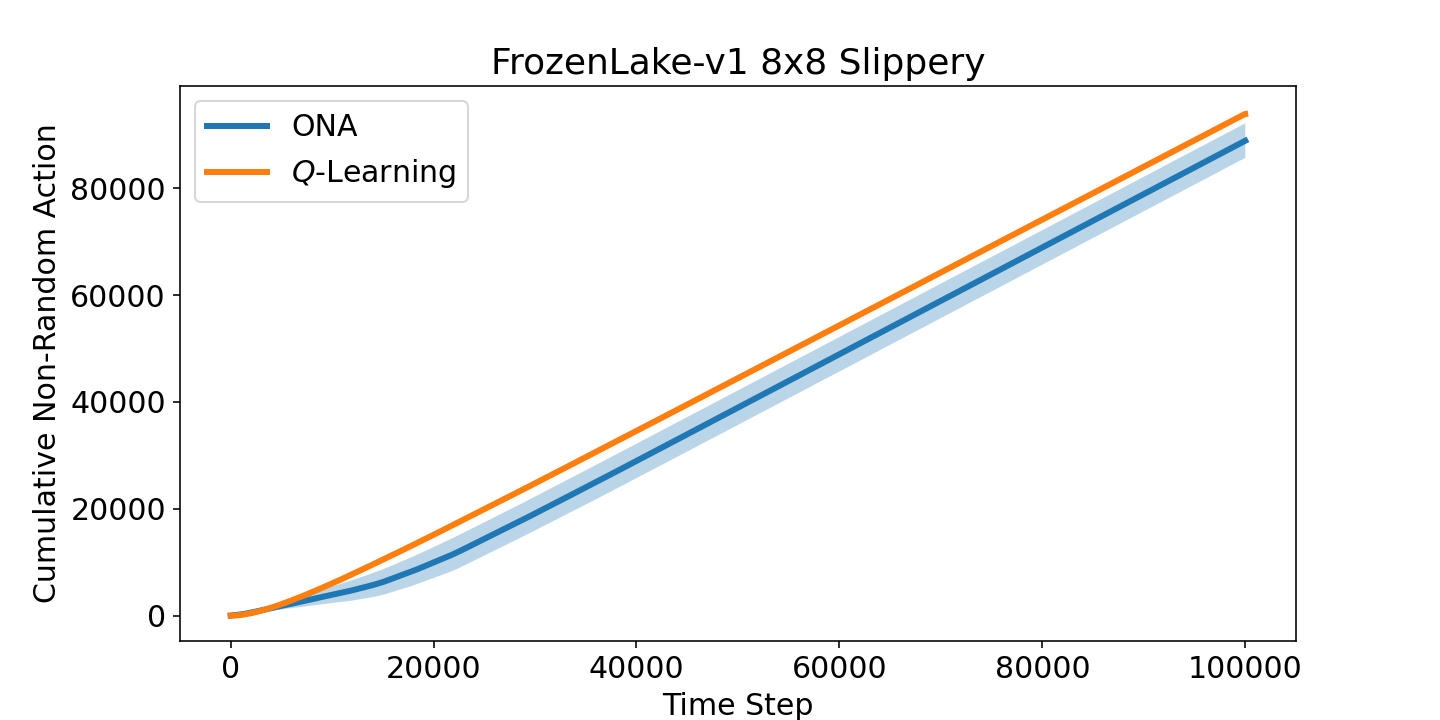}
         \caption{\scriptsize FrozenLake-v1 8x8 Slippery}
         \label{Cumulative_Non-Random_Action_vs_Time_Step_FrozenLake-v1_8x8_Slippery}
     \end{subfigure}
     \begin{subfigure}{0.32\columnwidth}
         \centering
         \includegraphics[width=\columnwidth]{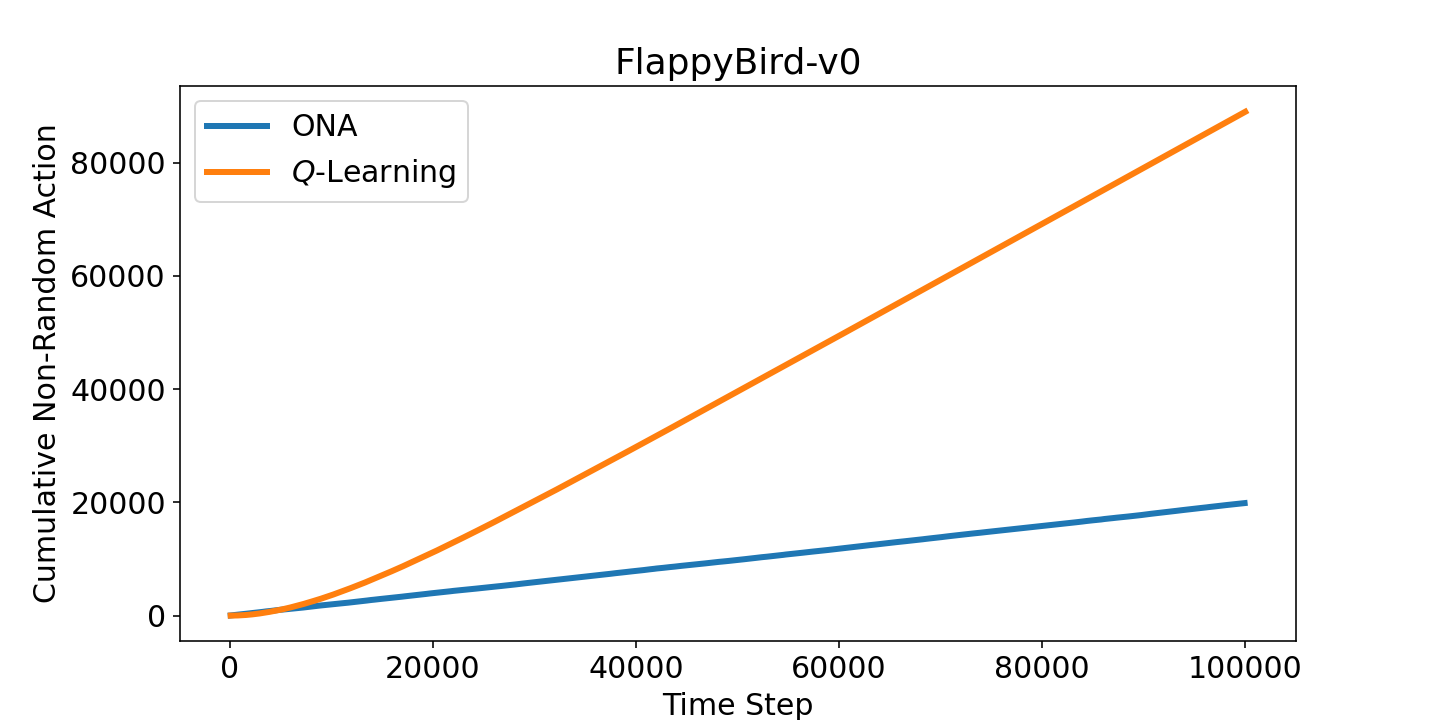}
         \caption{FlappyBird-v0}
         \label{Cumulative_Non-Random_Action_vs_Time_Step_FlappyBird-v0}
     \end{subfigure}
        \caption{Cumulative Non-Random Action vs. Time steps.}
        \label{Cumulative_Non-Random_Action_vs_Time Step}
\end{figure}
 
We also monitored the frequency of random action selection.
In Figures \ref{Cumulative_Random_Action_vs_Time Step} and \ref{Cumulative_Non-Random_Action_vs_Time Step}, the behavior of the two algorithms is drawn in terms of referring to a random or non-random action. In the case of $Q$-Learning, the probability of selecting a random action is primarily determined by the value of $\epsilon$. This probability decreases rapidly over time. Consequently, if the agent has not yet discovered a good policy or the environment changes, $Q$-Learning may not be able to solve the problem. Reducing $\epsilon$ over time can make it increasingly difficult to explore alternative solutions. The low variance of $Q$-Learning shows the agent's decision is unaffected by changes in the environment across different trials. This is because the random action-taking process is largely driven by $\epsilon$, which is independent of the problem.
On the other hand, in ONA, if the system does not suggest any action, we choose a random action, as shown in Figure \ref{Cumulative_Random_Action_vs_Time Step}.
In addition, there is also a possibility of the system suggesting a random action itself to explore the environment, thanks to the \textit{motorbabbling} parameter. 
So some of the actions shown in Figure \ref{Cumulative_Non-Random_Action_vs_Time Step}, despite being labeled as non-random, are actually exploratory and random in nature. 
In fact, ONA is able to decrease the \textit{motorbabbling} on its own once it has established stable hypotheses and accurate predictions, without relying on a reduction of the exploration rate that is dependent on time.
This could be one of the reasons why ONA is successful, even though it doesn't receive frequent rewards like $Q$-Learning. So, we believe ONA is capable of handling multiple and changing objectives, while also requiring less implicit example-dependent parameter tuning compared to $Q$-Learning.

\section{Conclusion and Future Work} \label{sectionC}
In this paper, we made a comparison between ONA and $Q$-learning on seven tasks. 
Given that both approaches demonstrate comparable performance on average, our study suggests that NARS has the potential to be a viable substitute for RL in sequence-based tasks, particularly for non-deterministic problems.
While further research is necessary to determine the full extent of NARS's capabilities and limitations, our results offer new avenues for exploration and innovation in the field of machine learning.
Future work can extend our examples to multi-objective and changing objective scenarios. Additionally, a combination of both approaches through methods like voting, hierarchical or teacher-student learning could be explored.

\section*{Acknowledgment}
We would like to express our gratitude to Patrick Hammer, Ph.D., for his expert advice, encouragement, and proofreading of the manuscript throughout this work.
This work was partially supported by the Swedish Research Council through grant agreement no. 2020-03607 and in part by Digital Futures, the C3.ai Digital Transformation Institute, and Sweden's Innovation Agency (Vinnova). 
The computations were enabled by resources in project SNIC 2022/22-942 provided by the Swedish National Infrastructure for Computing (SNIC) at Chalmers Centre for Computational Science and Engineering (C3SE).
\bibliographystyle{splncs04}
\bibliography{AGI-Project-without_url.bib}

\begin{thebibliography}{10}
\providecommand{\url}[1]{\texttt{#1}}
\providecommand{\urlprefix}{URL }
\providecommand{\doi}[1]{https://doi.org/#1}

\bibitem{brockman2016openai}
Brockman, G., Cheung, V., Pettersson, L., Schneider, J., Schulman, J., Tang,
  J., Zaremba, W.: Openai gym. arXiv preprint arXiv:1606.01540  (2016)

\bibitem{eberding2020sage}
Eberding, L.M., Th{\'o}risson, K.R., Sheikhlar, A., Andrason, S.P.: Sage:
  task-environment platform for evaluating a broad range of ai learners. In:
  International Conference on Artificial General Intelligence. pp. 72--82.
  Springer (2020)

\bibitem{fischer2018}
Fischer, T.G.: Reinforcement learning in financial markets-a survey. Tech.
  rep., FAU Discussion Papers in Economics (2018)

\bibitem{hammer2021autonomy}
Hammer, P.: Autonomy through real-time learning and OpenNARS for Applications.
  Temple University (2021)

\bibitem{hammer2020opennars}
Hammer, P., Lofthouse, T.: ‘opennars for applications’: architecture and
  control. In: International Conference on Artificial General Intelligence. pp.
  193--204. Springer (2020)

\bibitem{hammer2020reasoning}
Hammer, P., Lofthouse, T., Fenoglio, E., Latapie, H., Wang, P.: A reasoning
  based model for anomaly detection in the smart city domain. In: Proceedings
  of SAI Intelligent Systems Conference. pp. 144--159. Springer (2020)

\bibitem{hammer2016opennars}
Hammer, P., Lofthouse, T., Wang, P.: The opennars implementation of the
  non-axiomatic reasoning system. In: International conference on artificial
  general intelligence. pp. 160--170. Springer (2016)

\bibitem{henderson2018deep}
Henderson, P., Islam, R., Bachman, P., Pineau, J., Precup, D., Meger, D.: Deep
  reinforcement learning that matters. In: Proceedings of the AAAI conference
  on artificial intelligence. vol.~32 (2018)

\bibitem{kober2013}
Kober, J., Bagnell, J.A., Peters, J.: Reinforcement learning in robotics: A
  survey. The International Journal of Robotics Research  \textbf{32}(11),
  1238--1274 (2013)

\bibitem{silver2017}
Silver, D., Schrittwieser, J., Simonyan, K., Antonoglou, I., Huang, A., Guez,
  A., Hubert, T., Baker, L., Lai, M., Bolton, A., et~al.: Mastering the game of
  go without human knowledge. nature  \textbf{550}(7676),  354--359 (2017)

\bibitem{sutton2018}
Sutton, R.S., Barto, A.G.: Reinforcement learning: An introduction. MIT press
  (2018)

\bibitem{wang1995non}
Wang, P.: Non-axiomatic reasoning system: Exploring the essence of
  intelligence. Indiana University (1995)

\bibitem{wang2006rigid}
Wang, P.: Rigid flexibility: The logic of intelligence, vol.~34. Springer
  Science \& Business Media (2006)

\bibitem{wang2009insufficient}
Wang, P.: Insufficient knowledge and resources—a biological constraint and
  its functional implications. In: 2009 AAAI Fall Symposium Series (2009)

\bibitem{wang2010non}
Wang, P.: Non-axiomatic logic (nal) specification. University of Camerino,
  Piazza Cavour  \textbf{19} (2010)

\bibitem{wang2013non}
Wang, P.: Non-axiomatic logic: A model of intelligent reasoning. World
  Scientific (2013)

\bibitem{watkins1992q}
Watkins, C.J., Dayan, P.: Q-learning. Machine learning  \textbf{8}(3),
  279--292 (1992)

\bibitem{10.1145/3477600}
Yu, C., Liu, J., Nemati, S., Yin, G.: Reinforcement learning in healthcare: A
  survey. ACM Comput. Surv.  \textbf{55}(1) (nov 2021)

\end{thebibliography}

\end{sloppypar}
\end{document}